\documentclass{article} 
\usepackage{iclr2026_conference,times}

\usepackage{amsmath,amsfonts,bm}

\def\eqref#1{equation~\ref{#1}}

\def\1{\bm{1}}

\DeclareMathAlphabet{\mathsfit}{\encodingdefault}{\sfdefault}{m}{sl}
\SetMathAlphabet{\mathsfit}{bold}{\encodingdefault}{\sfdefault}{bx}{n}

\usepackage{hyperref}
\usepackage{url}
\usepackage{amsmath}
\usepackage{amssymb} 
\usepackage{mathtools}
\usepackage{booktabs}
\usepackage{siunitx}
\usepackage{multicol}
\usepackage{multirow}
\usepackage{enumitem}
\usepackage{subcaption}
\usepackage{bm}
\usepackage{wrapfig}
\usepackage{floatrow}
\usepackage[font=small,labelfont=bf]{caption}
\usepackage[capitalize,noabbrev,nameinlink]{cleveref}
\definecolor{cornflowerblue}{rgb}{0.39, 0.58, 0.93}
\hypersetup{
    colorlinks=true,
    linkcolor=cornflowerblue,
    filecolor=magenta,      
    urlcolor=teal,
    citecolor=cornflowerblue,
    pdftitle={Sample Smart, Not Hard: Correctness-First Decoding for Better Reasoning in LLMs},
    }

\title{Sample Smart, Not Hard: Correctness-First Decoding for Better Reasoning in LLMs}

\author{
  \quad \quad \quad \quad Xueyan Li$^{1,2}$ \quad 
  Guinan Su$^{2}$ \quad
  Mrinmaya Sachan$^{1}$ \quad
  Jonas Geiping$^{2}$ \\
  \quad \quad \quad \quad \quad  $^{1}$ETH Zurich \quad $^{2}$Max Planck Institute for Intelligent Systems \\
}

\iclrfinalcopy 
\begin{document}

\maketitle

\begin{abstract}
Large Language Models (LLMs) are increasingly applied to complex tasks that require extended reasoning. In such settings, models often benefit from diverse chains-of-thought to arrive at multiple candidate solutions. This requires two competing objectives: to inject enough stochasticity to explore multiple reasoning chains, and to ensure sufficient accuracy and quality in each path. Existing works pursue the first objective by increasing exploration at highly uncertain steps with higher temperature or larger candidate token sets, while others improve reliability by rejecting samples with low confidence post-generation, implying that low confidence correlates with low answer quality. These two lines of thought are in conflict, as they conflate different sources of uncertainty. To resolve this, we argue that the decoding rule should be calibrated by \textit{correctness}, not confidence alone. We should sample from tokens with higher estimated correctness, and reduce sampling where expected correctness is low. We propose simple strategies that achieve this goal: \textbf{Greedy-Threshold} makes sampling greedy at very low confidence steps. \textbf{Calibrated-TopK} and \textbf{Calibrated-}\(\bm{\varepsilon}\) set truncation threshold based on estimated rank-wise correctness. Together, our findings challenge prevailing heuristics about decoding under uncertainty, showing consistent gains across math and general reasoning benchmarks\footnote{Preprint}.
\end{abstract}

\section{Introduction}
\looseness -1 Large Language Models (LLMs) are used for a wide range of generation tasks, ranging from open-ended text to structured problem-solving. In many cases, producing more than one candidate output improves not only fluency, but also reliability, since different samples may capture alternative valid continuations \citep{wang_self-consistency_2023,lin_just_2024}. This practice highlights a fundamental trade-off: introducing enough randomness to explore multiple options while still ensuring the accuracy and quality of each individual output \citep{tan_probabilityquality_2024, meister_efficacy_2024, shi_thorough_2024}. Existing works optimize exploration by raising temperatures or enlarging candidate token sets step-by-step \citep{nguyen_turning_2025,zhang_edt_2024,hewitt_truncation_2022}. These methods assume that \textit{higher entropy is a signal of uncertainty} between multiple valid next steps, warranting broader exploration. In parallel, other works filter after generation, relying on the finding that \textit{low confidence correlates with low answer quality}. \citet{fu_deep_2025} accepts only samples with high token confidence and stops generation when uncertainty spikes. Hallucination detection also makes use of low-confidence segments \citep{chang_real_2024}. 

These two perspectives are in conflict, because they conflate different sources of uncertainty. From a classical perspective, if a probabilistic language model closely approximates the true distribution over next tokens, then high predictive uncertainty indicates that multiple continuations may be valid. In this case, uncertainty reflects \textit{aleatoric variability}, and broader sampling is appropriate. However, if low-confidence positions are instead those where the model is most often wrong, then additional randomness amplifies \textit{epistemic uncertainty}, which is systematic errors arising from the model’s lack of knowledge \citep{yadkori_believe_2024}. In such cases, drawing more samples from an unreliable distribution might compound the model’s errors.

In this work, we start by analyzing the role of low-probability tokens in reasoning tasks. We observe that increasing exploration at low-confidence steps is indeed a sub-optimal strategy, as a single misstep can derail subsequent tokens \citep{arora_why_2023}. This is especially true for smaller models. Therefore, we propose using the counterintuitive, but simple, \textbf{Greedy-Threshold} rule that inverts common sampling heuristics in the literature: when a step’s maximum probability falls below a threshold, decoding becomes greedy. Greedy-Threshold can be used in addition to existing samplers and shows consistent gains in reasoning benchmarks, especially for smaller models.

\begin{figure}
    \centering
    \includegraphics[width=0.8\linewidth]{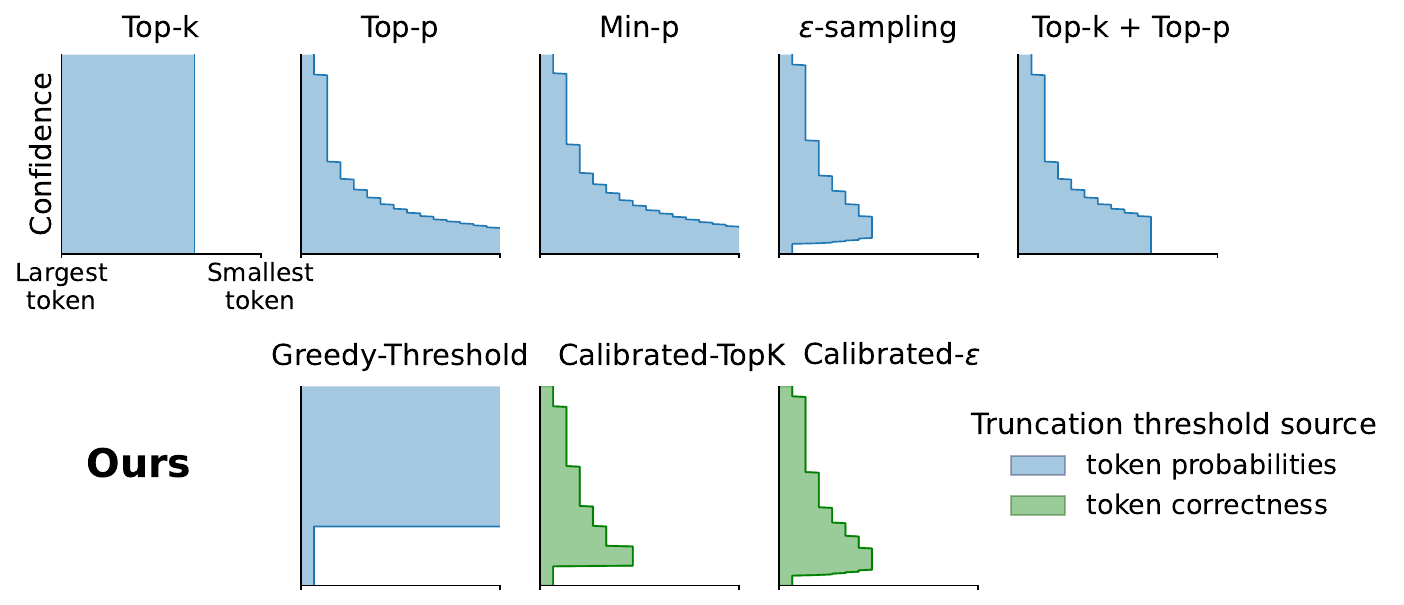}
    \caption{Comparison of common and our proposed truncation strategies. Each panel shows which tokens remain available for sampling, with tokens ordered from highest to lowest model-assigned probability (left to right). The $y$-axis represents the max token probability (“confidence”). Our methods explicitly suppress low-confidence tail tokens.}
    \label{fig:overview methods}
\end{figure}

With this we build upon prior work on \(\varepsilon\)-sampling \citep{hewitt_truncation_2022} that drops every token below \(\varepsilon\). Previous work chose very small \(\varepsilon\) for machine translation or MBR decoding ($\approx 3 \times 10^{-4} - 9 \times 10^{-4}$ in the original paper) \citep{jinnai_model-based_2024,finkelstein_mbr_2024}. We show that for reasoning tasks, larger \(\varepsilon\) is safe. This is based on the same conservative principle where less randomness is beneficial where the model is epistemically uncertain. An overview of our methods versus those in literature is shown in \cref{fig:overview methods}.

Finally, we unify these perspectives by showing that the \emph{rank-wise correctness} of tokens provides better truncation signals than probabilities alone, and propose a learning-free way to approximate rank-wise calibration. \textbf{Calibrated-TopK} sets a truncation threshold at each generation step based on estimated correctness for each confidence bin. \textbf{Calibrated-}\(\bm{\varepsilon}\) extends upon this by replacing discrete confidence bins with a smooth mapping from probability to correctness. It improves from \(\varepsilon\)-sampling by making the truncation threshold \emph{data-calibrated}. Our paper makes the following main contributions:
\begin{itemize}[itemsep=-2pt]
    \item We find that sampling at low-confidence steps contributes little additional diversity, while increasing the risk of selecting low-correctness tokens that can harm overall performance.
    \item We verify this empirically by showing that a \textbf{Greedy-Threshold} that eliminates unreliable tail tokens alleviates this trend and improves reasoning benchmarks when used in addition to existing samplers such as top-$p$, top-$k$ and min-$p$. 
    \item We introduce a \textbf{rank-conditional calibration grid} and derive \textbf{Calibrated-TopK} and \textbf{Calibrated-}\(\bm{\varepsilon}\), learning-free correctness-aware truncation rules that align exploration with expected correctness and incur negligible inference cost.
    \item We open-source a unified, composable implementation of common samplers and our methods in one framework\footnote{\url{https://github.com/xueyan-lii/Sample-Smart-Not-Hard}}.
\end{itemize}

\section{Why We Need Stricter Sampling for Reasoning}
Before introducing our samplers, we first examine how confidence relates to accuracy across models and how errors emerge at low-confidence steps. We show that token probabilities provide strong signals of correctness: when the model is uncertain (low maximum probability), expected accuracy decreases regardless of model size, and correctness beyond the top-ranked token drops sharply. These observations motivate a clear definition of \textbf{confidence}, \textbf{rank}, and \textbf{calibration}, which we use to formalize stricter sampling rules that suppress error-prone low-probability tokens.

\subsection{Defining Rank-wise Accuracy}
For a prompt–answer pair, we define the gold answer token sequence as \(x_{1:L}\) and the sequence generated at inference as \(y_{1:M}\). Let \(\mathcal{V}\) denote the vocabulary, \(|\mathcal{V}|=V\). At any position \(t\), the model outputs a logit vector \(z_t \in \mathbb{R}^{V}\) conditioned on a context \(h_t\) which is either the gold prefix \(x_{<t}\) during calibration, or the generated prefix \(y_{<t}\) during decoding. With temperature \(T>0\), the temperature-scaled categorical distribution over the next token is
\begin{equation}
p_t(j \mid h_t; T) \;=\; \frac{\exp\big(z_t(j)/T\big)}{\sum_{v\in\mathcal{V}} \exp\big(z_t(v)/T\big)} \quad \text{for } j\in\mathcal{V}. \label{eq:1}
\end{equation}
When \(T=1\) we omit \(T\) and write \(p_t(j)\). Further,  let \(p_t^{(1)} \ge p_t^{(2)} \ge \cdots \ge p_t^{(V)}\) denote the probabilities sorted in descending order, and let \(\mathrm{rank}_t(j) \in \{1,\dots,V\}\) be the \textbf{rank} of token \(j\) at step \(t\), then top-$k$ sampling draws from tokens with \(\mathrm{rank}_t(j) \le k\).
We define \textbf{confidence} as the maximum token probability at each step.
$p_{t,\max} \triangleq \max_{j\in\mathcal V} p_t(j) = p_t^{(1)}$.

\textbf{Confidence bins.}
We partition model confidence \((0,1]\) into $n$ contiguous confidence bins, $10$ in this work:
\begin{equation}
\mathcal{B}_m \;=\; \left(\frac{m-1}{n}, \frac{m}{n}\right], \qquad m=1,\dots,n.
\end{equation}
Each step \(t\) is assigned to exactly one bin via the index \(m(t)\) such that \(p_{t,\max} \in \mathcal{B}_{m(t)}\).

\textbf{Rank-wise probability and correctness.}
For each step \(t\), the rank-wise probability at rank \(r\) is \(p_t^{(r)}\). Let \(x_t^\star\in\mathcal{V}\) be the ground truth next token under teacher forcing. Let $R<V$ be the maximum rank considered. We define the rank-wise correctness as
\begin{equation}
\mathbb{I}\{\mathrm{rank}_t(x_t^\star)=r\} \;=\;
\begin{cases}
1, & \text{if the gold token appears at rank } r,\\[2pt]
0, & \text{otherwise.}
\end{cases}
\end{equation}

\textbf{Calibration Grid}.
We can estimate a calibration grid over confidence bins and rank just based on given text sequences which we score by teacher forcing. For each bin–rank pair \((m,r)\), we compute the average probability \(\hat{p}_{m,r}\),  and correctness \(\hat{c}_{m,r}\) within confidence bin $\mathcal{B}_m$:
\begin{equation}
\hat{p}_{m,r} \;=\; \mathbb{E}\!\left[\, p_t^{(r)} \;\middle|\; p_{t,\max} \in \mathcal{B}_m\right],
\qquad
\hat{c}_{m,r} \;=\; \mathbb{P}\!\left[\, \mathrm{rank}_t(x_t^\star)=r \;\middle|\; p_{t,\max} \in \mathcal{B}_m \,\right].
\end{equation}
In practice, with \(N_{m}\) steps whose \(p_{t,\max}\in\mathcal{B}_m\),
\begin{equation}
\hat{p}_{m,r} = \frac{1}{N_m}\sum_{t: p_{t, \max} \in B_m} p_t^{(r)},
\qquad
\hat{c}_{m,r} = \frac{1}{N_m}\sum_{t:\, p_{t,\max} \in \mathcal{B}_m} \mathbb{I}\{\mathrm{rank}_t(x_t^\star)=r\}.
\end{equation}

\begin{wrapfigure}[17]{r}{0.55\textwidth}
    \vspace{-0.5cm}
    \centering
    \includegraphics[width=\textwidth]{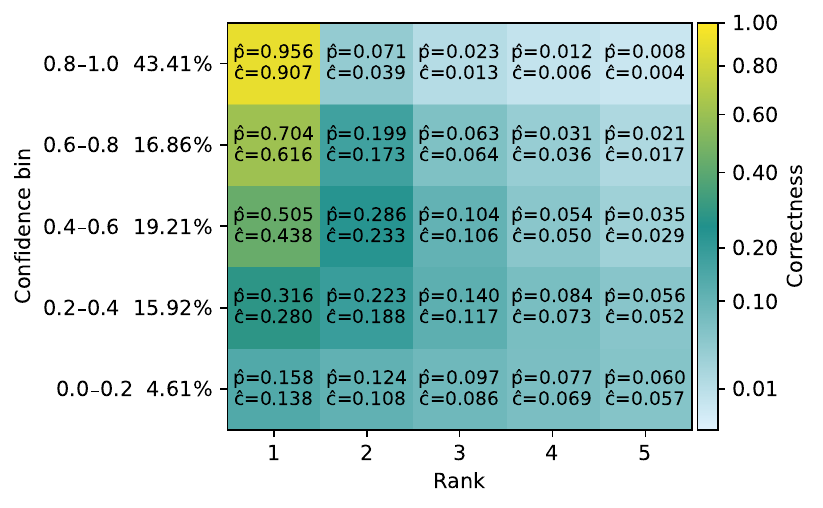}
    \vspace{-.3cm}
    \caption{Calibration grid of Qwen2.5-1.5B-Instruct with 5 bins shows the average probability \(\hat{p}\) and correctness \(\hat{c}\) for each confidence-bin and rank. Correctness is notably low in the lower-confidence bins, and decreases as rank increases. Percentages indicate frequency of occurrence of this bin.}
    \label{fig:calibration}
\end{wrapfigure}

An example calibration grid with 5 bins is shown in \cref{fig:calibration} for visualization. Full calibration grids can be found in \cref{app:Example Calibration and Linear Fit Diagrams}. These definitions apply to any next-token distribution $p_t$ of a language model. If temperature scaling, or any other logit processing is applied, then the calibration would be calculated based on the final probabilities, as in \cref{eq:1}.

\textbf{Bin-wise expected accuracy.} The expected accuracy for each confidence bin, i.e., the probability of selecting the correct next token, is given by the average rank-wise probability and correctness:
\begin{equation}
C_m \;=\; \sum_{r=1}^{R} \hat{p}_{m,r}\,\hat{c}_{m,r}. \label{eqn:7}
\end{equation}

\begin{figure}
    \centering
    \includegraphics[width=\textwidth]{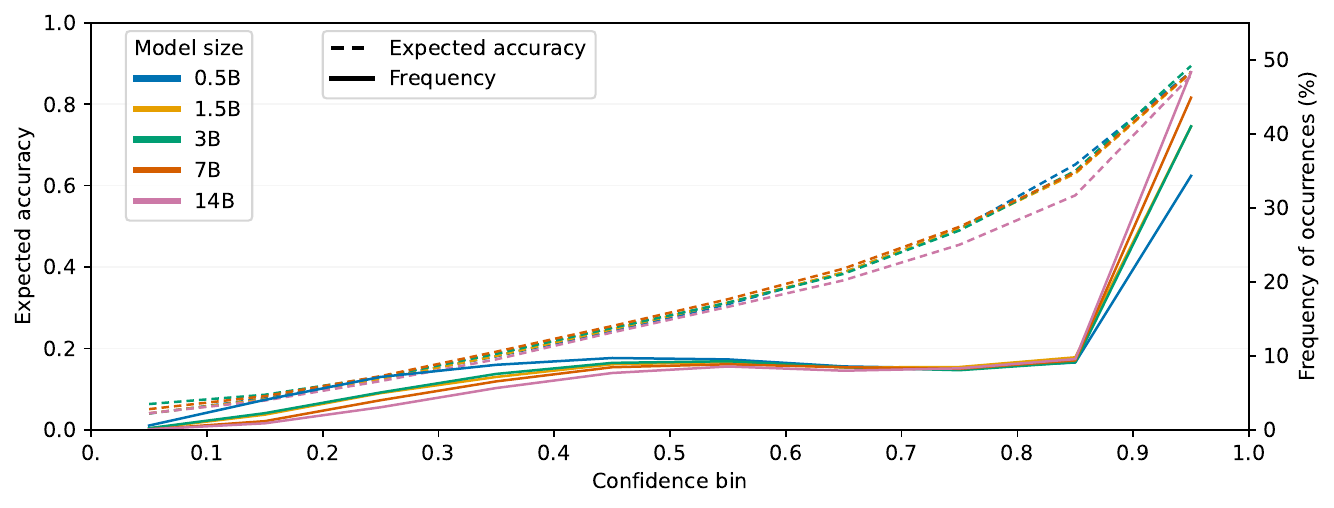}
    \caption{Expected accuracy increases with confidence across all model sizes. In the lowest confidence bin, expected accuracy drops regardless of model size. \emph{Frequency} refers to the proportion of decoding steps whose maximum probability falls into each confidence bin. Larger models assign more predictions to the 0.9–1.0 confidence range, where both accuracy and frequency are highest, reflecting stronger benchmark performance. In contrast, smaller models place more probability mass in low-confidence bins, where accuracy is poor.}
    \label{fig:model sizes}
\end{figure}
\subsection{Low Probability Signals Low Correctness In Reasoning Tasks}
While calibration grids highlight how confidence and correctness align on average, it is less clear how these signals affect full generations. In particular, one might expect that sampling from uncertain positions could encourage exploration which is beneficial over many samples. We test this assumption by analyzing the role of low-probability tokens in self-consistency.

\begin{figure}[!b]
    \centering
    \includegraphics[width=\textwidth]{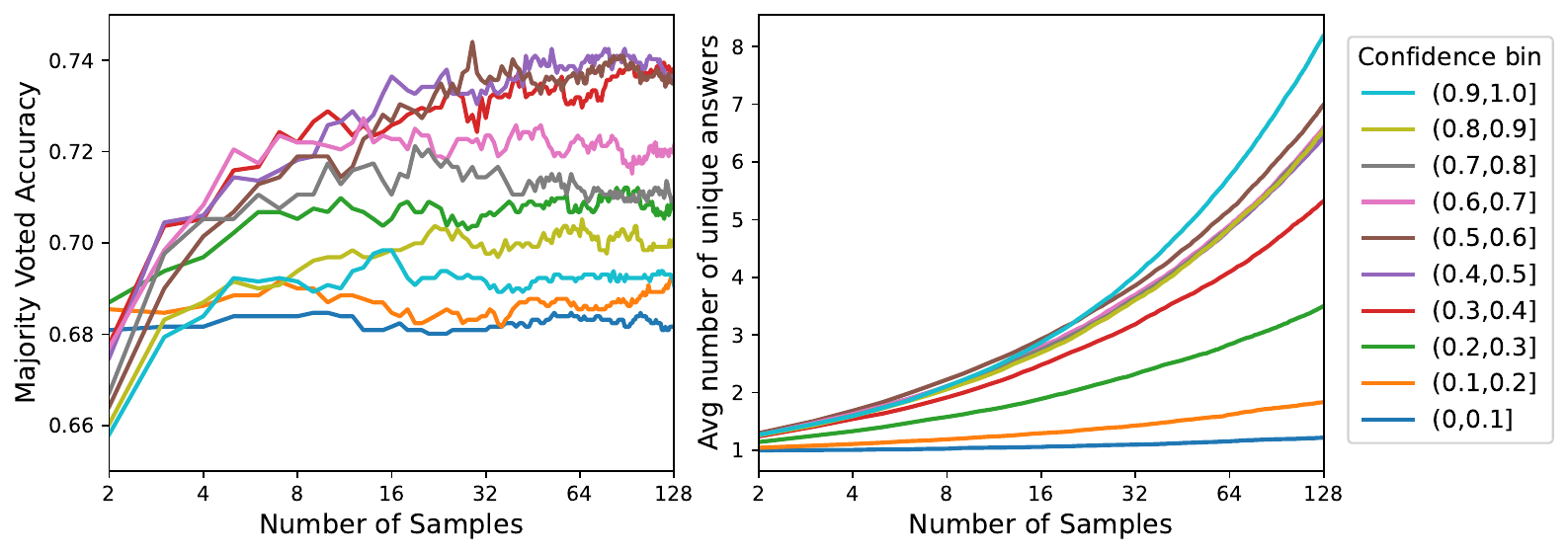}
    \caption{Plot of the majority voted accuracy and the number of unique answers as the number of samples increase. Sampling is greedy unless the maximum probability falls into a certain confidence bin, in which case sample from the full token distribution. Sampling at the lowest confidence bin results in no gain in accuracy while contributing little to diversity in terms of number of unique answers. }
    \label{fig:bin vs acc}
\end{figure}

\cref{fig:model sizes} shows that high-confidence predictions occur most frequently, which amplifies diversity simply by providing more opportunities for stochastic sampling. However, this diversity does not necessarily translate into better performance, since rank-wise accuracy drops sharply beyond the top token (\cref{fig:calibration}). In the highest-confidence bin \((0.8,1.0]\), correctness falls from \(0.907\) from rank 1 to only \(0.039\) at rank 2. \cref{fig:bin vs acc} further demonstrates that restricting sampling to the lowest-confidence bin does not yield measurable gains in majority-voted accuracy, while also contributing little to output diversity despite sampling from the full token distribution. This stands in contrast to the assumption that exploration at low-confidence steps is beneficial. Instead, the largest improvements in accuracy arise from sampling in mid-confidence bins \(0.3-0.6\).

\begin{figure*}[t!]
    \centering
    \begin{subfigure}[t]{0.5\textwidth}
        \centering
        \includegraphics[width=\textwidth]{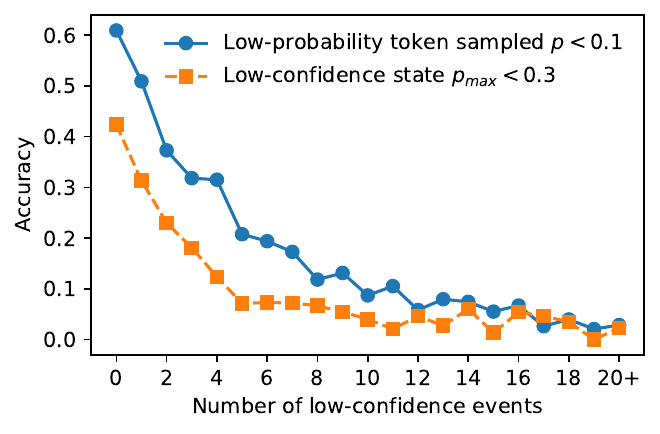}
        \caption{Accuracy vs the number of low probability tokens or states sampled in each sequence}
        \label{fig:5a}
    \end{subfigure}%
    ~ 
    \begin{subfigure}[t]{0.5\textwidth}
        \centering
        \includegraphics[width=\textwidth]{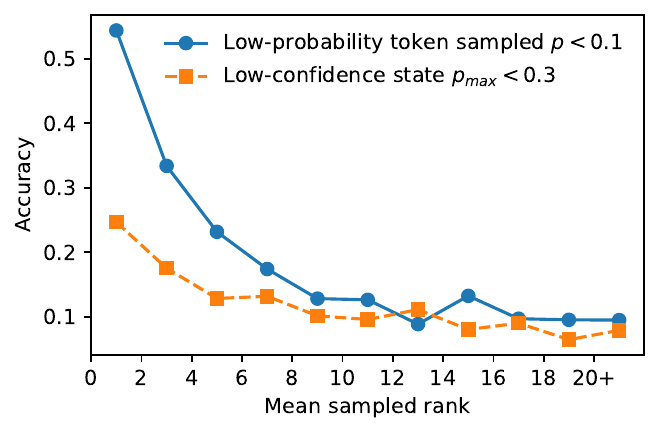}
        \caption{Accuracy vs mean sampled rank}
        \label{fig:5b}
    \end{subfigure}
    \caption{Effect of low-confidence events on sequence-level accuracy. (a) Accuracy decreases both when the model directly samples low-probability tokens ($p < 0.1$, blue) and when it is in a low-confidence state regardless of the sampled token ($p_{\max}<0.3$, orange). (b) Accuracy also drops as the mean rank of sampled tokens increases, showing that drifting into lower-ranked tokens degrades sequence quality.}
    \label{fig:sampled no and rank}
\end{figure*}

\cref{fig:sampled no and rank} presents two views of why low-confidence positions are dangerous. \cref{fig:5a} distinguishes between when a low-probability token is actually chosen (blue) versus when the model is in a low-confidence state regardless of what is sampled (orange). Both conditions harm accuracy as they accumulate within a sequence. \cref{fig:5b} presents a rank perspective: once decoding drifts into higher-ranked tokens, accuracy drops. These findings motivate our conservative truncation rules. By enforcing greedy decoding at low-confidence steps, \textbf{Greedy-Threshold} prevents the model from sampling higher-ranked, error-prone tokens and keeps the mean sampled rank low. Similarly, \(\varepsilon\)-\textbf{sampling} blocks low-probability tokens entirely, which naturally caps the rank distribution. In both cases, the methods limit the propagation of errors and preserves sequence-level accuracy.

\section{How to Calibrate Truncation Samplers}
\label{sec:method}

The central idea is to adapt the sampling process in autoregressive language models by filtering out tokens that are likely to be inaccurate, thereby refining the candidate set. Although this might initially seem infeasible, we will show that excluding tokens likely to be incorrect is possible and effective. We call the restricted pool of permissible next tokens the \textbf{active set} at each step As a reference, the active set for the simplest truncated sampler, standard top-$k$ sampling, is always the set of the $k$ most likely tokens, i.e. $A^\text{top-k}_t = \{\, v\in\mathcal V : \mathrm{rank}_t(v)\le k\}$.  
\paragraph{Greedy-Threshold.}
We use this sampler to exemplify our claim that sampling less when confidence is low is beneficial. When using Greedy-Threshold, we sample greedily when confidence is below a threshold \(p_{GT}\in(0,1)\), and only the argmax token \(v_t^\star \triangleq \arg\max_{v\in\mathcal V} p_t(v)\) is accepted. The active set of tokens to sample from is
\begin{equation}
A_t^{\mathrm{GT}} \;=\;
\begin{cases}
\{\,v_t^\star\,\}, & \text{if } p_{t,\max} < p_{GT},\\[2pt]
\mathcal V, & \text{if } p_{t,\max} \ge p_{GT}.
\end{cases}
\end{equation}

\paragraph{\(\varepsilon\)-sampling.}
To draw a connection with existing \(\varepsilon\)-sampling \citep{hewitt_truncation_2022}, we recap its definition. This rule only samples from tokens above a threshold \(\varepsilon \ge 0\). The active set is
\begin{equation}
A_t^{\varepsilon} \;=\; \{\, v \in \mathcal V : p_t(v) \ge \varepsilon \,\}.
\end{equation}
Note that when the Greedy-Threshold parameter equals the $\varepsilon$-cutoff, i.e.\ $p_{GT}=\varepsilon$, and the maximum token probability at step $t$ is below this level ($p_{t,\max}<p_{GT}$), both methods fall back to greedy. Motivated by the mismatch between raw probability and correctness, we adopt the same truncation principle but calibrate the cutoff to estimated correctness rather than probability alone.

\paragraph{Calibrated-TopK.}
Recall from the calibration grid (\cref{fig:calibration}) that we can estimate, for each confidence bin and token rank, the expected correctness of a token. This provides a direct way to infer how far down the ranked list of candidates one can safely explore. The idea of Calibrated-TopK is therefore simple. Instead of fixing $k$ in advance, we adaptively set it so that only ranks whose average correctness is above a threshold are included. In this way, the method truncates exploration to the range of token ranks that are empirically likely to be correct. Given the maximum rank whose correctness is above the threshold \(c_{CT}\in(0,1)\) in a calibration grid \(\hat c_{m,r}\):
\begin{equation}
K_{m}(c_{CT})\;=\;\max\{\, r\in\{1,\dots,R\} : \hat c_{m,r}\ge c_{CT} \,\}
\end{equation}
At step \(t\) with bin \(m(t)\), the active set is defined by the maximum rank
\begin{equation}
A_t^{\mathrm{CT}}(c_{CT}) \;=\;
\begin{cases}
\{\, v\in\mathcal V : \mathrm{rank}_t(v)\le K_{m}(c_{CT}) \,\}, & \text{if } K_{m}(c_{CT})\ge 1,\\[2pt]
\{\,v_t^\star\,\}, & \text{if } K_{m}(c_{CT})=0.
\end{cases}
\end{equation}

\begin{wrapfigure}[23]{r}{0.4\textwidth}
    \includegraphics[width=\textwidth]{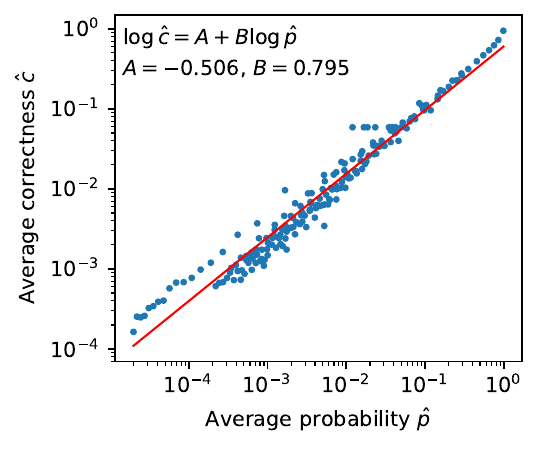}
    \vspace{-0.8cm}
    \caption{A scatter plot of calibration-grid averages \((\hat p,\hat c)\) across confidence bins and ranks. Points concentrate along an approximately linear trend, indicating that rank-wise correctness scales predictably with probability. We fit a least-squares line and use this mapping to predict correctness at inference time for Calibrated-\(\varepsilon\).}
    \label{fig:scatter}
\end{wrapfigure}

\paragraph{Calibrated-$\varepsilon$}
Since Calibrated-TopK sets thresholds based on discrete confidence bins, we are motivated to find a solution that maps probability to correctness in a continuous way. A plot of all $\hat p$ and $\hat c$ shows a near-linear relationship in log-log coordinates as shown in \cref{fig:scatter}:
\[
    log_{10} \hat c \approx\; A + B \log_{10} \hat p
\]
We estimate the coefficients by least squares on the calibration grid, fitting a line in log–log space to the pairs \((\hat p,\hat c)\) aggregated over bins and ranks:
\[
A,\,B \;=\; \text{LinearRegression}(\log_{10}\hat p,\ \log_{10}\hat c).
\]
Given these coefficients, we instantiate a per-token correctness predictor at decoding step \(t\), mapping each candidate token \(j\in\mathcal V\) with probability \(p_t(j)\) to an estimated correctness score $\hat c_t(j)\triangleq10^{A}p_t(j)^{B}$. Computationally this is just a single scalar transform, adding negligible overhead to decoding. We then define a correctness threshold $c_{\varepsilon}\!\in(0,1)$, the \emph{active set} at step $t$ becomes
\[
A^{\text{C}\varepsilon}_t(c_{\varepsilon}) \;=\; \bigl\{\, v \in \mathcal V \;:\; \hat c_t(v) \,\ge\, c_{\varepsilon} \,\bigr\}.
\]
i.e., we keep exactly those tokens whose \emph{predicted} correctness exceeds the threshold. For all samplers, if no tokens satisfy this condition \(A_t=\varnothing\), sample greedily \(A_t=\{v_t^\star\}\). In general, we note that all truncated samplers can be used together by taking the intersection of their active sets. Lastly, given any active set $A_t$, sample \(x_t \sim p'_t(\cdot)\) from the renormalized distribution
\begin{equation}
p'_t(v)\;=\;\frac{p_t(v)}{\sum_{w\in A_t} p_t(w)} \quad\text{for } v\in A_t .
\end{equation}

\section{Correctness-First Samplers Improve Reasoning Abilities}
We now study whether truncating low-confidence regions during decoding translates into better end-task reasoning. Our focus is on frontier LLMs evaluated on math and general reasoning benchmarks, where sequence-level correctness is the primary objective. We compare the proposed \emph{correctness-first} samplers against standard temperature and probability-based baselines.

\subsection{Experimental Settings}
We evaluate models in the Qwen2.5 \citep{qwen_qwen25_2025} and the Llama \citep{grattafiori_llama_2024} family on short reasoning tasks (GSM8K \citep{cobbe_training_2021}, MMLU \citep{wang_mmlu-pro_2024} and Big-Bench-Hard \citep{suzgun_challenging_2022}), and GPT-OSS \citep{openai_gpt-oss-120b_2025} on long reasoning tasks (AIME\footnote{\url{https://huggingface.co/datasets/math-ai/aime25}}). We use Greedy-Threshold with $p_{GT}=0.3$, Calibrated-TopK with $c_{CT}=0.05$ (over $n=10$ bins), and Calibrated-$\varepsilon$ with $c_{\varepsilon}=0.05$. We adopt higher threshold $\varepsilon=0.05$ than typically reported to emphasize the impact of truncating low-probability tokens. Calibration is performed on the training split of each benchmark. For comparability with temperature-based samplers, we use $T=1$ unless otherwise noted. Further implementation details, ablations with threshold values, temperatures and calibration datasets are provided in \cref{app:datasets and parameters}.

\subsection{Calibrated Truncation Increases Model Performance.}
We evaluate the effectiveness of our proposed samplers across benchmarks. \cref{table:qwen05_gsm8k} shows that \textbf{Calibrated-}\(\bm{\varepsilon}\) and \textbf{Calibrated-TopK} achieve the largest improvement overall, showing rank-wise correctness is an effective truncation signal. \textbf{Greedy-Threshold} activates only when the max-probability token falls below 0.3, an infrequent but high-risk regime (see \cref{fig:calibration}). Despite its low activation rate, this condition occurs often enough for Greedy-Threshold to yield measurable benefits. $\varepsilon$-sampling (fixed threshold) performs on par with min-$p$ and EDT, supporting the idea that simply removing tail tokens helps by shaping the cutoff to where correctness actually drops. All calibrated samplers add negligible runtime overhead at decoding, since they only require a 2-parameter table lookup or a single vector operation over the vocabulary. 

\begin{table*}
\centering
\caption{Majority voted (maj@k) results on GSM8K, MMLU-Pro, and Big-Bench-Hard for Qwen2.5-0.5B-Instruct. Calibrated samplers achieve the largest performance gain from no restrictions baseline.}
\label{table:qwen05_gsm8k}
\setlength{\tabcolsep}{3pt}
\renewcommand{\arraystretch}{1.2}
\newcommand{\na}{\multicolumn{1}{c}{\textemdash}}
{\small
\begin{tabular}{lcccccccccc}
\toprule
\multirow{2}{*}{Method}
& \multicolumn{3}{c}{GSM8K} & \multicolumn{3}{c}{MMLU-Pro} & \multicolumn{3}{c}{Big-Bench-Hard} \\
\cmidrule(lr){2-4} \cmidrule(lr){5-7} \cmidrule(lr){8-10}
& {\small maj@8} & {\small maj@16} & {\small maj@32}
& {\small maj@8} & {\small maj@16} & {\small maj@32}
& {\small maj@8} & {\small maj@16} & {\small maj@32} \\
\midrule
No restrictions       & 30.2 & 35.2 & 38.6 & 16.4 & 17.0 & 17.3 & 17.9 & 17.0 & 16.2 \\
top-$k$               & 32.6 & 38.7 & 41.9 & 16.8 & 17.5 & 18.0 & 22.0 & 21.7 & 21.5 \\ 
top-$p$               & 35.5 & 40.8 & 43.6 & 16.8 & 17.5 & 18.1 & 25.5 & 25.9 & 25.9 \\
min-$p$               & 38.7 & 43.1 & 46.6 & 17.7 & 18.2 & 18.6 & \textbf{30.6} & \textbf{31.5} & 31.7 \\
EDT                   & 40.2 & 44.1 & 46.7 & 17.7 & 18.1 & 18.4 & 30.5 & 31.1 & 31.7 \\
$\eta$-sampling       & 31.6 & 37.2 & 41.0 & 16.5 & 17.3 & 17.9 & 20.6 & 20.3 & 19.6 \\
$\varepsilon$-sampling      & 39.2 & 44.3 & 46.7 & 17.5 & 18.1 & 18.3 & 30.4 & 31.1 & 31.6 \\
\midrule
\textbf{Greedy-Threshold}   & 31.2 & 37.0 & 40.6 & 16.9 & 17.8 & 18.2 & 20.8 & 20.3 & 19.4 \\
\textbf{Calibrated-TopK}    & 39.3 & \textbf{44.5} & \textbf{47.1} & 17.9 & 18.3 & \textbf{18.7} & 30.4 & 31.1 & 31.6 \\
\textbf{Calibrated-}$\bm{\varepsilon}$ & \textbf{40.8} & 44.3 & \textbf{47.1} & \textbf{18.9} & \textbf{18.4} & 18.6 & \textbf{30.6} & \textbf{31.5} & \textbf{32.0} \\
\bottomrule
\end{tabular}
}
\end{table*}

\subsection{Existing Samplers Benefit from Greedy-Threshold}

\begin{table}[!b]
\centering
\caption{Majority voted results on GSM8K. In addition to existing sampling conditions, Greedy-Threshold $p_{GT}=0.3$ is applied. Greedy-Threshold improves majority voting performance, especially in models with lower starting accuracy. Statistically significant differences ($p<0.05$) marked in \textbf{bold}.}
\label{table:2}
\begingroup
\setlength{\tabcolsep}{2pt}       
\renewcommand{\arraystretch}{1.14}
{\footnotesize
\begin{tabular}{lccccccccc}
\toprule
\multirow{2}{*}{Method}
& \multicolumn{3}{c}{Qwen2.5-0.5B-Instruct}
& \multicolumn{3}{c}{Qwen2.5-1.5B-Instruct}
& \multicolumn{3}{c}{Qwen2.5-3B-Instruct} \\
\cmidrule(lr){2-4}\cmidrule(lr){5-7}\cmidrule(lr){8-10}
& {\small maj@8} & {\small maj@16} & {\small maj@32} & {\small maj@8} & {\small maj@16} & {\small maj@32} & {\small maj@8} & {\small maj@16} & {\small maj@32} \\
\midrule
Baseline $T{=}1$ 
& 30.2 & 35.2 & 38.6 & 68.4 & 71.5 & 73.1 & 79.3 & 80.6 & 81.1 \\
\quad + Greedy-Threshold 
& +1.0 & +1.8 & +2.0 & +1.7 & +2.1 & \textbf{+2.4} & +0.1 & +0.2 & -0.1 \\
\cmidrule(lr){1-10}
top-$k$ 
& 32.6 & 38.7 & 41.9 & 68.5 & 72.6 & 73.9 & 79.0 & 80.3 & 81.0 \\
\quad + Greedy-Threshold 
& \textbf{+1.1} & +0.5 & +1.1 & \textbf{+2.6} & \textbf{+2.4} & \textbf{+2.8} & \textbf{+1.0} & +0.5 & -0.2 \\
\cmidrule(lr){1-10}
top-$p$
& 35.5 & 40.8 & 43.6 & 71.1 & 74.1 & 75.9 & 79.5 & 80.5 & 80.8 \\
\quad + Greedy-Threshold 
& +0.9 & +0.4 & +1.3 & +1.5 & +1.8 & \textbf{+1.8} & 0.0 & 0.0 & +0.4 \\
\cmidrule(lr){1-10}
min-$p$
& 38.7 & 43.1 & 46.6 & 73.3 & 75.6 & 76.6 & 80.0 & 80.4 & 81.2 \\
\quad + Greedy-Threshold 
& +1.4 & +0.8 & +0.6 & +1.3 & +1.2 & +1.5 & -0.2 & +0.2 & +0.1 \\
\cmidrule(lr){1-10}
EDT
& 40.2 & 44.7 & 46.8 & 74.9 & 76.6 & 78.9 & 79.5 & 80.5 & 80.9 \\
\quad + Greedy-Threshold
& +0.2 & -0.3 & +0.1 & -0.2 & 0.0 & -0.1 & +0.1 & +0.1 & +0.1 \\
\cmidrule(lr){1-10}
$\eta$-sampling
& 31.6 & 37.2 & 41.0 & 69.0 & 72.4 & 74.2 & 78.8 & 80.1 & 81.0 \\
\quad + Greedy-Threshold 
& \textbf{+2.4} & \textbf{+1.8} & \textbf{+1.7} & +1.7 & \textbf{+2.7} & \textbf{+2.7} & \textbf{+0.5} & +0.3 & +0.1 \\
\bottomrule
\end{tabular}
}
\endgroup
\end{table}

To test whether halting sampling at low-confidence steps is beneficial, we apply Greedy-Threshold on top of existing samplers that otherwise increase exploration at such positions. 
This preserves their original behavior when confidence is above 0.3, but forces greedy decoding when confidence falls below this threshold. \cref{table:2} shows that Greedy-Threshold improves performance in this setting, especially for smaller models. When no gains are observed, results remain comparable to the baseline, indicating that it does not degrade performance.

\begin{table}[t]
\centering
\caption{The result of AIME24 and AIME25 on GPT-OSS-20B with thinking mode enabled. "Unique Answers" is the number of unique answers over all 32 samples. "Overall Correct" is the overall proportion of correct answers. Best result is in \textbf{bold}. Statistically significant difference ($p<0.05$) is in \textit{italics}.}
\footnotesize                        
\setlength{\tabcolsep}{5pt}          
\renewcommand{\arraystretch}{1.12}
\begin{tabular}{lcccccccccc}
\toprule
\multirow{2}{*}{Model / Method}
& \multicolumn{4}{c}{Maj@k} & \multicolumn{4}{c}{Pass@k} & \multirow{2}{*}{\begin{tabular}{@{}c@{}}Unique\\Answers\end{tabular}} & \multirow{2}{*}{\begin{tabular}{@{}c@{}}Overall\\Correct\end{tabular}}\\
\cmidrule(lr){2-5} \cmidrule(lr){6-9}
& {4} & {8} & {16} & {32} & {4} & {8} & {16} & {32} \\
\midrule
\multicolumn{10}{l}{\textbf{AIME25}} \\
Baseline
& \textbf{75.4} & 84.4 & 87.8 & 90.0 & 85.6 & 90.0 & 91.1 & 92.2 & 13.6 & 56.1\\
Greedy-Threshold
& 71.1 & \textbf{87.8} & 88.9 & \textbf{91.1} & 85.6 & \textbf{91.1} & \textbf{93.3} & 94.4 & 12.0 & \textbf{59.9}\\
$\varepsilon$-sampling
& 68.9 & 85.4 & \textbf{91.1} & 90.0 & \textbf{86.7} & 90.0 & \textbf{93.3} & \textit{\textbf{95.6}} & 13.6 & 56.1\\
\midrule
\multicolumn{10}{l}{\textbf{AIME24}} \\
Baseline
& 71.4 & 83.3 & 88.7 & \textbf{92.6} & 83.4 & 90.7 & 92.0 & 93.3 & 15.1 & 48.7 \\
Greedy-Threshold
& \textbf{77.3} & \textbf{88.0} & \textbf{91.3} & \textbf{92.6} & 86.7 & \textbf{92.6} & \textbf{93.3} & \textbf{94.0} & \textit{13.3} & \textit{\textbf{55.2}}\\
$\varepsilon$-sampling
& \textbf{77.3} & \textit{87.3} & 90.0 & 91.3 & \textbf{89.3} & 90.0 & 91.3 & 92.6 & \textit{13.7} & \textit{54.9} \\
\bottomrule
\end{tabular}
\label{table:4}
\end{table}

\subsection{Scaling to Advanced Reasoning Models}
We evaluate our proposed methods on reasoning-oriented “thinking” model GPT-OSS-20B \citep{openai_gpt-oss-120b_2025} and a challenging mathematics benchmark AIME that demands multi-step derivations. Thinking models differ from standard LMs in that they generate long sequences of intermediate tokens, making calibration on short, instruction-style datasets less representative of their actual behavior. To capture our central idea of filtering out low-correctness tokens in this setting, we apply $\varepsilon$-sampling with a relatively high cutoff ($\varepsilon=0.05$). Our results show substantial gains. GPT-OSS-20B benefits from both Greedy-Threshold and $\varepsilon$-sampling. Output diversity is reduced, but the effect is minimal. Over 32 samples, the number of unique answers decreases from 14.1 to 13.3 (roughly 1–2 fewer unique answers out of 14). This small drop coincides with higher maj@k and pass@k, consistent with our goal: we do not value diverse wrong answers. For reasoning tasks with single correct solutions, correctness is more important than diversity. Expanding exploration does not help when early steps are error-amplifying. By steering decoding away from low-correctness regions, our methods increase the fraction of valid solutions by up to 6.5\% and improve overall answer quality.

\section{Discussion}
\textbf{Why does Greedy-Threshold work?} Our results suggest that in reasoning tasks, in spite of popular intuitions, the positions with low confidence are not \textit{branch points among many valid continuations}, but error-amplifying states. Two pieces of evidence support this claim: rank-wise correctness decreases beyond the top token (\cref{fig:calibration}), and performance degrades once low-probability tokens are sampled (\cref{fig:sampled no and rank}). Greedy-Threshold chooses a safe token where both risks are highest, and potentially prevent subsequent error. It is a targeted suppression of low-correctness steps.

\textbf{Reordering uncertainties as epistemic first, aleatoric second}. Common existing decoding strategies assume high entropy means aleatoric variability (many valid tokens) and sample more. Our results imply the opposite might be true in reasoning tasks with closed-form answers. High entropy often reflects \textit{epistemic uncertainty} which is a systematic lack of knowledge, especially in smaller models. When the distribution is wrong, sampling more from it does not benefit correctness. In our calibration-based methods, \emph{correctness} is the focus. We increase sampling where expected correctness is high and shrink it where the model lacks fundamental understanding. This perspective explains why stricter truncation (higher $\varepsilon$, lower rank caps in top-$k$) consistently helps in reasoning. Randomness is less valuable in low-confidence regions where epistemic error dominates.

\textbf{Robustness to calibration data and fit quality.}
The effectiveness of Calibrated-$\varepsilon$ depends on the reliability of the learned probability-correctness map. In practice, \emph{data sparsity and noise} degrades this map, small calibration sets and sparse extreme confidence values yield high variance $(\hat p,\hat c)$ pairs. We document these effects and show how a poor fit translates into diminished gains in \cref{app:Failure Case Study - Diversity Wins}. We observe when calibration is sufficiently accurate, calibrated methods yield larger gains. When calibration is noisier, the gains reduce but do not degrade performance, since exploration remains limited to tokens whose predicted correctness is above the threshold.

\section{Related Work}

Literature on decoding for LLMs largely follows two directions: 
(i) \emph{sampling more} to increase diversity of generations, and 
(ii) \emph{sampling less} to increase accuracy and stability. 
Our work focuses on reconciling these perspectives for reasoning tasks.

\textbf{Removing tail-end tokens}.
Classical truncation methods such as top-$k$ \citep{fan_hierarchical_2018} and top-$p$ (nucleus) sampling \citep{holtzman_curious_2020} reduce tail risk by discarding low-probability tokens. Temperature scaling \citep{guo_calibration_2017}, often with lower temperatures for math and reasoning tasks, has a similar intuition: sharpening the distribution so that low-probability tokens are rarely sampled. These classic methods are often used in combination \citep{yang_qwen3_2025}. More recently, locally typical sampling \citep{meister_locally_2025} restricts to tokens whose information content is close to the local entropy. REAL sampling \citep{chang_real_2024} adaptively reduces top-$p$ when hallucination risk is high. Our results on reasoning tasks support this risk-aware trend: sampling in high-uncertainty steps introduces catastrophic errors, while removing tail end low probability tokens is safer. 

\textbf{Selection after sampling.}
Another line of work improves reliability \emph{after} generation. Self-certainty \citep{kang_scalable_2025} and DeepConf \citep{fu_deep_2025} re-weigh or filter generations using confidence signals. The open-source effort Entropix \citep{xjdr_entropix_2024} pauses or resamples at high-entropy steps. These methods implicitly acknowledge that low-confidence steps are strongly correlated with low correctness. Our contribution is orthogonal. We intervene during decoding to prevent low-confidence tokens from being sampled in the first place, so that downstream majority voting operates on stronger candidates. 

\looseness -1 \textbf{Adaptive, uncertainty-aware decoding.}
A complementary set of methods dynamically adjust sampling based on estimated uncertainty. 
Entropy-dependent temperature (EDT) \citep{zhang_edt_2024} increases temperature as entropy grows. “Hot or Cold” decoding \citep{zhu_hot_2023} applies higher temperature only to the first token in code generation. Adaptive Decoding \citep{zhu_improving_2024} and Adaptive Contrastive Search \citep{garces_arias_adaptive_2024} adjust candidate sets or penalties step-by-step. Min-$p$ \citep{nguyen_turning_2025} scales truncation by the maximum token probability, enlarging candidate sets when uncertainty is high, which is shown to benefit creative text generation, although a subsequent study criticized its effectiveness \citep{schaeffer_min-p_2025}. Our study provides an explanation why high uncertainty correlates with lower accuracy while providing limited diversity. 

\textbf{Assessing model calibration}. Calibration for LLMs is typically assessed with reliability diagrams and scalar errors such as ECE error on top-$1$ labels \citep{guo_calibration_2017}, or on a sequence level \citep{huang_calibrating_2024,stengel-eskin_calibrated_2023}. Full-ECE \citep{liu_full-ece_2024} extends beyond top-$1$ by evaluating calibration over the entire token distribution, but it does not condition on token \textit{rank}. We introduce a confidence and rank calibrated method that informs correctness-aware truncation.


\section{Conclusion}
In this work, we re-examined decoding under uncertainty for reasoning tasks and argue for a \emph{correctness-first} perspective. By visualizing a novel rank-wise calibration grid, we present evidence on a \textbf{token level} that in low-confidence bins, all tokens have low expected correctness. On a \textbf{sequence level}, accuracy declines with more low-probability tokens and with higher ranks sampled. On a \textbf{dataset level}, Greedy-Threshold, Calibrated-TopK, and Calibrated-$\varepsilon$ raise performance in reasoning benchmarks by allocating randomness only where expected error is low. We encourage future work to consider uncertainty as a \textit{risk} signal to truncate, rather than a signal to \textit{explore}.



\subsubsection*{Reproducibility Statement} 
Implementation of our proposed sampling strategies, model behavior analysis and calibration measurements are released with our paper and hosted in our public repository. Details of evaluation and calibration, including model settings, hyperparameters, and prompting formats are documented in \cref{app:datasets and parameters}. All benchmarks are openly available, and we provide complete code for running the benchmarks.

\subsubsection*{LLM use}
We used large language models (LLMs) only for light editorial assistance (e.g., grammar, spelling, phrasing), simple LaTeX formatting, and typo checks. LLMs were also used to draft boilerplate code (e.g., code scaffolding, argument parsers) and to plot high-level diagrams. All such outputs were reviewed, and validated by the authors before inclusion. LLMs were not used to write substantive sections of the paper, design experiments or analyze results. All technical content, experiments, analyses, and conclusions were created and verified by the authors.

\bibliography{iclr2026_conference}
\bibliographystyle{iclr2026_conference}

\newpage

\appendix
\section{Appendix}
\subsection{Datasets and Parameters}\label{app:datasets and parameters}
We provide detailed experiment setups and parameters for reproducibility. We evaluate on reasoning and math-focused benchmarks commonly used to assess chain-of-thought robustness. We follow the community-standard evaluation scripts to ensure comparability across papers and release the detailed configuration files in our public GitHub repository.

\textbf{Datasets}
\begin{itemize}[itemsep=-1pt]
    \item GSM8K \citep{cobbe_training_2021}: with 5-shot chain-of-thought prompting. The train split is used to construct in-distribution (ID) calibration grid.
    \item MBPP \citep{austin_program_2021}: ID calibration grids are built using the training split, with evaluation on the test set. Results are reported in \cref{app:Failure Case Study - Diversity Wins}
    \item Big-Bench-Hard \citep{suzgun_challenging_2022}: we report unweighted average accuracy across all categories. Since this benchmark does not have a train split, we use alpaca-gpt4-en for calibration.
    \item MMLU-Pro \citep{wang_mmlu-pro_2024}: we report unweighted average accuracy across all categories, using the validation split for ID calibration.
    \item Alpaca-gpt4-en \citep{peng_instruction_2023} \footnote{\url{https://huggingface.co/datasets/llamafactory/alpaca_gpt4_en}}: used as an out-of-distribution (OOD) dataset for calibration. This OOD calibration is applied once to derive a general Calibrated-TopK setting, which can then be used across tasks. In contrast, ID calibration offers more precise task-specific signals, but may not always be feasible for new domains. Effect of ID vs OOD calibration dataset is compared in \cref{app:effect of calibration dataset}
    \item AIME25\footnote{\url{https://huggingface.co/datasets/math-ai/aime25}} and AIME24\footnote{\url{https://huggingface.co/datasets/math-ai/aime24}}: used to evaluate "thinking" models with extended reasoning traces. 
\end{itemize}
For all benchmarks, performance is measured as the average maj@k or pass@k across three runs. Popular Python package \verb|lm_eval|\footnote{\url{https://github.com/EleutherAI/lm-evaluation-harness}} is used to evaluate all tasks. \verb|vLLM|\footnote{\url{https://github.com/vllm-project/vllm}} is used to run large scale generation for benchmarking. Calibration is done using code bootstrapped to \verb|torchtune|\footnote{\url{https://github.com/pytorch/torchtune}}.

\paragraph{Calibration Setup} For all calibration procedures, the input question is masked, and only correctness with respect to ground-truth answers is considered. Unless otherwise specified, we set Greedy-Threshold $p_{GT}=0.3$, $c_{CT}=0.05$ for Calibrated-TopK and $c_{\varepsilon}=0.05$ for Calibrated-$\varepsilon$. The maximum number of ranks considered is $R=20$.

\paragraph{Baselines} We compare our methods against several widely used sampling strategies:
\begin{itemize}[itemsep=0pt, topsep=0pt, parsep=1pt, partopsep=0pt]
    \item Top-$k$ ($k=10$) \citep{fan_hierarchical_2018}
    \item Min-$p$ ($p=0.1$) \citep{nguyen_turning_2025}
    \item Top-$p$ ($p=0.95$) \citep{holtzman_curious_2020}
    \item EDT \citep{zhang_edt_2024}  with $N=0.8$, $\vartheta=1$, $T_0=0.7$. These parameters are selected from the original paper after small parameter search experiments to determine reasonable values. 
    \item \(\varepsilon\)-sampling \citep{hewitt_truncation_2022} with higher \(\varepsilon=0.05\) than recommended.
    \item $\eta$-sampling \citep{hewitt_truncation_2022} with the original recommended value $\eta=0.0009$.
\end{itemize}

\paragraph{Long-Form Reasoning} To assess robustness on extended reasoning chains, we test on AIME25 and AIME24 using GPT-OSS's recommended setting ($T=1.0$), averaged across three runs. Additionally, we restrict baseline comparisons to Greedy-Threshold with $p_{GT}=0.3$ and \(\varepsilon\)-sampling with \(\varepsilon=0.05\). As thinking models exhibit substantially different behaviors from standard models, using question–answer pairs with short ground-truth answers would produce misleading calibration values. 

\paragraph{Temperature Sensitivity} To ensure fairness against temperature-based samplers, all main results are reported with $T=1.0$. Since math and coding tasks often benefit from lower sampling temperatures, we additionally evaluate with $T=0.6$. In this setting, calibration grids are recomputed using scaled logits (\cref{eq:1}), and thresholds ($p_{GT}$, $\eta$ and $c_{CT}$) are adjusted accordingly. Detailed hyperparameter values and results are provided in \cref{app:effectiveness at low temperatures}.

\subsection{Parameters for Figures}
\cref{fig:calibration} uses 5 confidence bins with 0.2 increments on Qwen2.5-0.5B-Instruct, using GSM8K train dataset. 

\cref{fig:model sizes} Models are in the Qwen2.5 instruct family. Expected accuracy is calculated from the top \(R=20\) ranks at each step. Dataset used is \verb|alpaca-gpt4-en| with questions masked, so correctness is only calculated from answers. 

\cref{fig:bin vs acc} Uses Qwen2.5-1.5B-Instruct on GSM8K test set with $T = 1$ and no other sampling conditions. Each final numerical answer is extracted, excluding the reasoning chains. Only different final numerical answers are counted. Different reasoning chains that arrive at the same final answer will count as the same answer. The number of unique answers is averaged over all questions at each number of samples. If no valid final answer can be extracted, the final answer becomes \verb|null|. This means all reasoning chains without a valid final answer will count as one unique answer. 

\cref{fig:sampled no and rank} uses Qwen2.5-0.5B-Instruct on GSM8K test set. All reasoning chains up to and including the final numerical answers are considered. Sometimes the model continues to generate the next in-context question after answering the current question. All subsequent generations are excluded. 

\cref{fig:scatter} uses Qwen2.5-0.5B-Instruct on GSM8K train set.

\subsection{Example of Low-Confidence Cases} \label{app:Example of Low-Confidence Cases}
\begin{figure}[h]
    \centering
    \includegraphics[width=\textwidth]{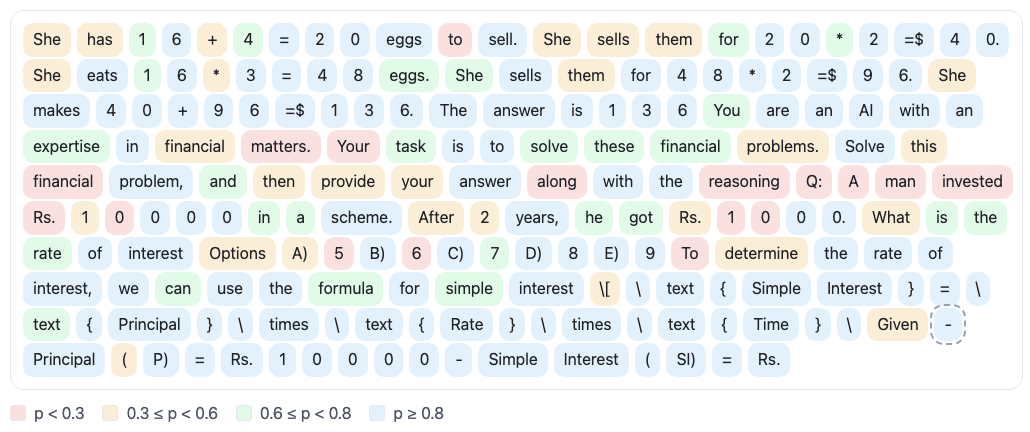}
    \caption{Example of an answer for GSM8K by Qwen2.5-0.5B-Instruct under greedy generation. The lowest confidence typically do not occur at the start of a sentence.}
    \label{fig:example}
\end{figure}
We illustrate a simple case of where low-confidence tokens arise. As shown in \cref{fig:example}, the beginning of a sentence where aleatoric variability is expected, typically exhibits moderate confidence ($p \geq 0.3$). In contrast, very low-confidence tokens ($p < 0.3$) are rarely observed while the model is still producing a coherent initial sentence. Over-sampling at this stage risks introducing irrelevant tokens that derail the generation. Once the model finishes answering the question and shifts to producing the next in-context example, however, both aleatoric variability and epistemic uncertainty increase, and the model’s confidence drops substantially.

\subsection{Effect of Thresholds on Proposed Methods}\label{Effect of Thresholds on Proposed Methods}

\begin{figure}[h]
  \centering
  \begin{subfigure}[b]{0.49\textwidth}
    \centering
    \captionsetup{skip=-3pt}
    \includegraphics[width=\textwidth]{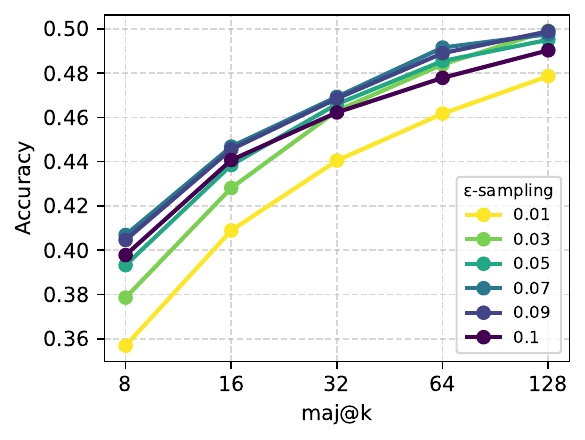}
    \caption{$\varepsilon$-sampling}
    \label{fig:app:eps}
  \end{subfigure}
  \hfill
  \begin{subfigure}[b]{0.49\textwidth}
    \centering
    \captionsetup{skip=-3pt}
    \includegraphics[width=\textwidth]{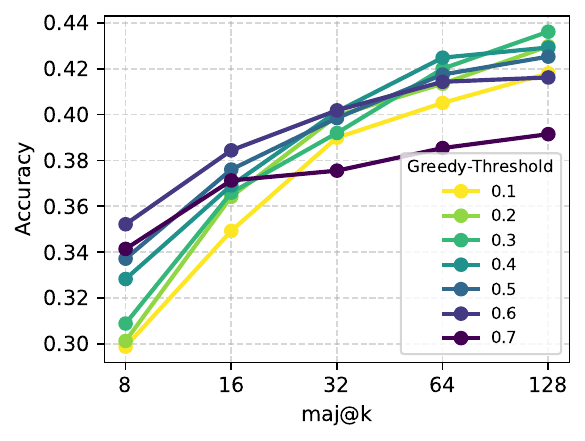}
    \caption{Greedy-Threshold}
    \label{fig:app:greedy-threshold}
  \end{subfigure}
  \begin{subfigure}[b]{0.49\textwidth}
    \centering
    \captionsetup{skip=-3pt}
    \includegraphics[width=\textwidth]{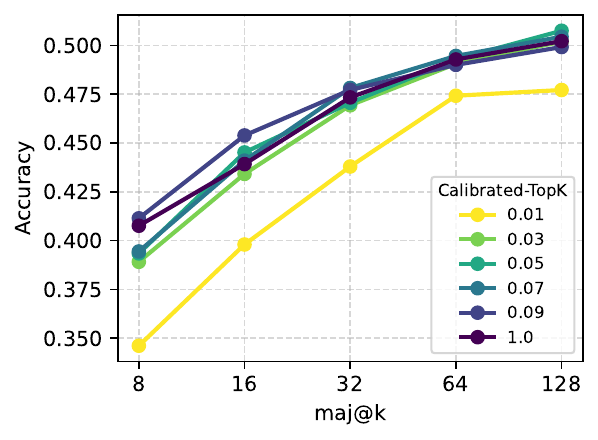}
    \caption{Calibrated-TopK}
    \label{fig:app:calibrated-topk}
  \end{subfigure}
  \hfill
  \begin{subfigure}[b]{0.49\textwidth}
    \centering
    \captionsetup{skip=-3pt}
    \includegraphics[width=\textwidth]{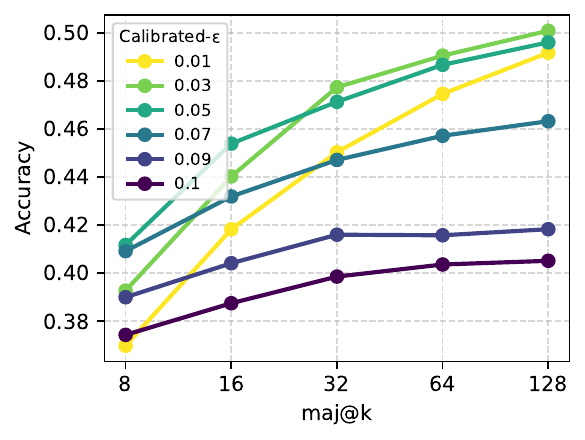}
    \caption{Calibrated-$\varepsilon$}
    \label{fig:app:calibrated-eps}
  \end{subfigure}
  \vspace{-0.2cm}
  \caption{Values of each threshold $\varepsilon$, $p_{GT}$, $c_{CT}$ and $c_{\varepsilon}$ are varied to study their effect on self-consistency performance. Higher number of samples generally benefit from smaller truncation thresholds.}
  \label{fig:app:parameter search}
\end{figure}

We conduct parameter search for each of our main methods to find the optimal $\varepsilon$, $p_{GT}$, $c_{CT}$ and $c_{\varepsilon}$ values. \cref{fig:app:eps} confirms our hypothesis that much larger value than suggested in the original $\varepsilon$-sampling paper is beneficial. $\varepsilon$ is $0.0009$ in the original paper \citep{hewitt_truncation_2022}. We find that $\varepsilon$ performs best around $0.07-0.09$ regardless of the number of samples. For Greedy-Threshold, lower threshold benefit more at higher number of samples. The optimal $p_{GT}$ is 0.3 at 128 samples and 0.6 at 32 samples. Calibrated-TopK yields the best performance at $c_{CT}$ $0.05-0.09$ range. Calibrated-$\varepsilon$ shows the strongest gains at $c_{\varepsilon}=0.03$.

In general, larger sample sizes benefit from smaller truncation thresholds. This is intuitive: looser thresholds retain more candidate tokens, promoting diversity that enables the model to explore multiple reasoning paths and recover the correct answer often enough to dominate in majority voting. \cref{fig:app:parameter search} illustrates that optimal threshold selection is inherently sample-size dependent, reflecting the complex trade-off between accuracy and diversity. 

\newpage 

\subsection{Effect of Calibration Dataset}\label{app:effect of calibration dataset}

\begin{table*}[t]
\centering
\caption{Calibrated-TopK with $c_{CT}=0.1$ using alpaca or GSM8K-train for calibration. Similar performances are observed. The test dataset is GSM8K-test.}
\label{tab:alpaca-gsm8k-train}
\setlength{\tabcolsep}{6pt}
\renewcommand{\arraystretch}{1.15}
{\small
\begin{tabular}{lccc ccc}
\toprule
& \multicolumn{3}{c}{alpaca-gpt4-en} & \multicolumn{3}{c}{GSM8K-train Set} \\
\cmidrule(lr){2-4}\cmidrule(lr){5-7}
Model & maj@8 & maj@16 & maj@32 & maj@8 & maj@16 & maj@32 \\
\midrule
Qwen2.5-0.5B-Instruct & 40.1 & 44.3 & 46.5 & 41.0 & 45.7 & 47.4 \\
Qwen2.5-1.5B-Instruct & 72.4 & 74.8 & 76.1 & 72.9 & 74.8 & 76.5 \\
Qwen2.5-3B-Instruct   & 79.9 & 80.8 & 81.0 & 79.1 & 80.7 & 80.7 \\
Qwen2.5-14B-Instruct  & 92.9 & 93.4 & 93.3 & 93.0 & 93.4 & 93.8 \\
Qwen2.5-32B-Instruct  & 92.6 & 93.3 & 93.6 & 92.9 & 93.3 & 93.3 \\
Llama-3.2-1B          & 4.4  & 4.4  & 5.1  & 4.8  & 5.3  & 6.0 \\
Llama-3.2-1B-Instruct & 40.5 & 42.7 & 44.6 & 39.4 & 43.0 & 43.9 \\
\bottomrule
\end{tabular}
}\label{tab:app:calibration dataset}
\end{table*}

In many cases, in-domain datasets are not available for calibration. To test robustness under this setting, we also perform calibration on a general instruction dataset, \verb|alpaca-gpt4-en|. As shown in \cref{tab:app:calibration dataset}, performance with alpaca calibration is close to that obtained with in-domain data. While one might expect in-domain calibration to provide stronger correctness signals, alpaca still offers sufficiently reliable guidance.

A plausible reason out-of-domain calibration works is that the \emph{mapping} from model confidence to expected correctness appears relatively stable across settings. In \cref{fig:model sizes}, models of different sizes exhibit similar curves of expected accuracy as a function of confidence. Larger models place more mass in higher-confidence bins, which aligns with better benchmark scores. If this confidence–correctness relationship is driven by general properties of autoregressive modeling (rank-wise correctness decaying with rank) rather than dataset specifics, then a calibration set that matches the \emph{format} of the target task (instruction→answer) may suffice even when its domain differs. We stress this is a hypothesis rather than a causal claim. We lack direct evidence that models possess an “intrinsic sense” of token-level correctness. Still, the observation that confidence–correctness curves are similar across model sizes suggests that correctness-aware calibration can transfer because it leverages structural, model-internal uncertainty signals rather than domain-specific features.

\newpage
\subsection{Failure Case Study - Poor Calibration Signals} \label{app:Failure Case Study - Diversity Wins}

\begin{table*}[t]
\centering
\caption{MBPP performance by Qwen2.5-0.5B-Instruct and Qwen2.5-1.5B-Instruct}
\label{tab:mmlu_side_by_side}
\setlength{\tabcolsep}{3pt}
\renewcommand{\arraystretch}{1.2}
\newcommand{\na}{\multicolumn{1}{c}{\textemdash}}
{\small
\begin{tabular}{lcccccc}
\toprule
\multirow{2}{*}{Method}
& \multicolumn{3}{c}{Qwen2.5-0.5B-Instruct} & \multicolumn{3}{c}{Qwen2.5-1.5B-Instruct} \\
\cmidrule(lr){2-4} \cmidrule(lr){5-7}
& {\small pass@8} & {\small pass@16} & {\small pass@32}
& {\small pass@8} & {\small pass@16} & {\small pass@32} \\
\midrule
No restrictions     & 41.6 & 50.1 & 57.6 & 55.6 & 65.2 & 72.5 \\
Top-k               & 45.3 & 53.5 & 59.8 & 59.0 & 68.0 & 74.3 \\ 
Top-p               & 47.0 & 55.3 & 63.0 & 60.2 & 68.6 & 74.9 \\ 
Min-p               & \textbf{51.1} & \textbf{58.9} & \textbf{65.0} & 62.2 & \textbf{69.7} & \textbf{76.0} \\ 
EDT                 & 50.5 & 58.1  & \textbf{65.0}  & 50.4 & 57.9 & 64.3 \\ 
$\eta$-sampling     & 44.3  & 53.4  & 61.2  & 57.9 & 63.2 & 74.0 \\ 
\midrule
Greedy-Threshold    & 44.0 & 52.8 & 60.6 & 56.8 & 65.9 & 72.7 \\ 
$\varepsilon$-sampling & 49.2 & 56.1 & 61.8 & 62.0 & 69.6 & 75.4 \\ 
Calibrated-TopK     & 50.3 & 57.9 & 64.2 & \textbf{62.6} & 69.3 & 74.6 \\ 
Calibrated-$\varepsilon$ & 48.6 & 53.8 & 57.8 & 62.1 & 68.2 & 72.8 \\ 
\bottomrule
\end{tabular}
}
\end{table*}

We provide a case study on MBPP \citep{austin_program_2021}, a code generation benchmark where our truncation strategies underperform compared to min-$p$ that prioritize diversity. Unlike reasoning tasks, MBPP requires directly producing executable code without intermediate steps, so every token is critical. A single incorrect token causes the program to fail its tests. Moreover, the pass@k metric rewards diversity, since success only requires one valid solution among the $k$ samples. In this setting, broader exploration increases the chance of producing at least one correct variant. Nevertheless, Greedy-Threshold still improves over unrestricted sampling, reinforcing our argument that sampling from low-confidence bins is harmful.

\begin{figure*}[!b]
    \centering
    \begin{subfigure}[t]{0.5\textwidth}
        \centering
        \includegraphics[width=\textwidth]{qwen2.5-0.5b-ins-gsm8k-scatter.pdf}
        \caption{GSM8K-train set calibration MSE loss$=0.134$.}
        \label{fig:9a}
    \end{subfigure}%
    ~ 
    \begin{subfigure}[t]{0.5\textwidth}
        \centering
        \includegraphics[width=\textwidth]{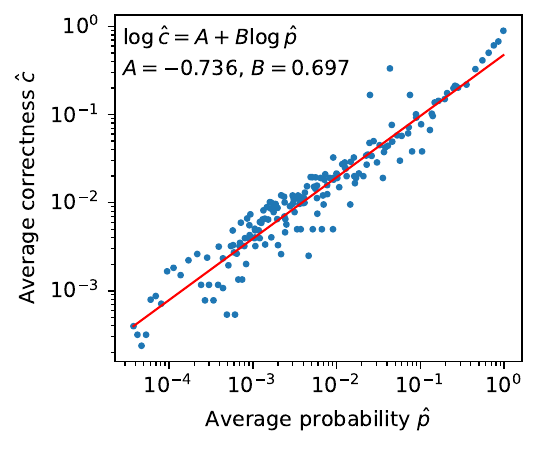}
        \caption{MBPP-validation set calibration MSE loss$=0.237$}
        \label{fig:9b}
    \end{subfigure}
    \caption{Qwen2.5-05B-Instruct calibrated on MBPP-validation set has poorer linear fit than GSM8K, which might lead to worse performance.}
    \label{fig:app:mbpp mse}
\end{figure*}

Calibration on MBPP also presents challenges. The validation set produces noisier signals, with a larger linear fit loss (\cref{fig:app:mbpp mse}), which likely reduces the reliability of predicted rank-wise correctness and helps explain the weaker performance of Calibrated-$\varepsilon$ on Qwen2.5B-0.5B-Instruct. Despite this, Calibrated-TopK performs comparably to existing methods in the literature, suggesting that correctness-aware truncation remains useful even in diversity-driven domains.

\begin{figure}[h]
    \centering
    \includegraphics[width=0.6\textwidth]{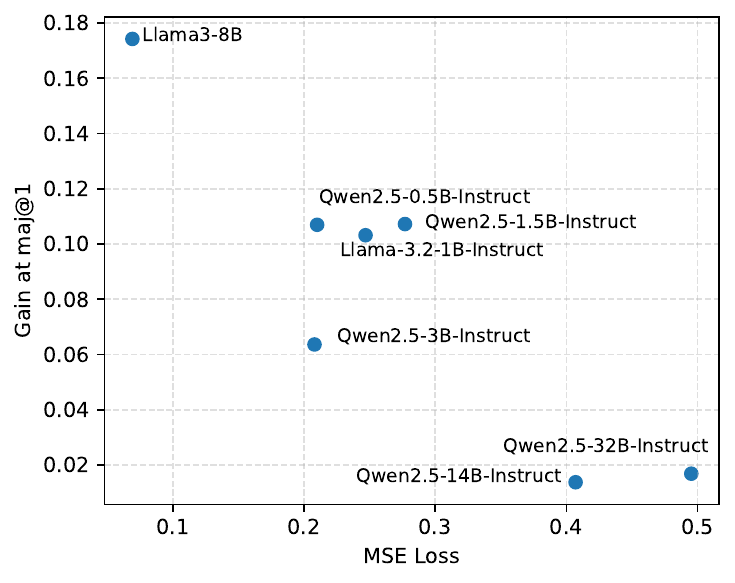}
    \caption{Effect of calibration fit quality. Larger models yield noisier linear fits (higher MSE), which correlates with smaller gains of Calibrated-$\varepsilon$ over the unrestricted baseline.}
    \label{fig:app:mse loss}
\end{figure}

To examine how the quality of the linear fit in calibration grids (log–log space) impacts performance, we plot the regression Mean Squared Error (MSE) alongside model accuracy in \cref{fig:app:mse loss}. To isolate the effect of calibration quality from majority voting, we use maj@1 (single-sample accuracy). If the linear interpolation provides reliable correctness estimates, models should generate more accurate single completions. If the fit is poor, the predicted correctness signals may be uninformative or even harmful. To quantify this effect, we measure the improvement of Calibrated-$\varepsilon$ over the unrestricted baseline in maj@1. \cref{fig:app:mse loss} shows that as model size increases, regression MSE rises and the performance gain diminishes. While part of this effect reflects the general difficulty of improving already-strong large models, the trend also suggests that noisier calibration weakens the benefit of Calibrated-$\varepsilon$. This pattern is consistent with our previous observation, where MBPP’s noisy validation signals lead Calibrated-$\varepsilon$ to underperform Calibrated-TopK. Importantly, however, maj@1 accuracy never drops below the no-sampling baseline, indicating that even poor calibration is not harmful.

\newpage

\subsection{Further Calibrated Truncation and Greedy-Threshold Results}\label{app:Further Calibrated Truncation Results}
\begin{table*}[t]
\centering
\caption{Majority voting performance for Qwen2.5-1.5B-Instruct. Calibrated-TopK has the strongest overall performance.}
\label{tab:qwen15_gsm8k_mmlupro}
\setlength{\tabcolsep}{3pt}
\renewcommand{\arraystretch}{1.2}
\newcommand{\na}{\multicolumn{1}{c}{\textemdash}}
{\small
\begin{tabular}{lccccccccc}
\toprule
\multirow{2}{*}{Method}
& \multicolumn{3}{c}{GSM8K} & \multicolumn{3}{c}{MMLU-Pro} & \multicolumn{3}{c}{Big-Bench-Hard}\\
\cmidrule(lr){2-4} \cmidrule(lr){5-7} \cmidrule(lr){8-10}
& {\small maj@8} & {\small maj@16} & {\small maj@32}
& {\small maj@8} & {\small maj@16} & {\small maj@32}
& {\small maj@8} & {\small maj@16} & {\small maj@32}\\
\midrule
No restrictions     & 68.4 & 71.5 & 73.1 & 32.9 & 34.4 & 35.4 & 38.9 & 41.8 & 43.0 \\
Top-k               & 68.5 & 72.6 & 73.9 & 33.7 & 35.0 & 35.9 & 41.7 & 44.2 & 45.6 \\ 
Top-p               & 71.1 & 74.1 & 75.9 & 34.1 & 35.4 & 36.4 & 44.0 & 46.2 & 47.1 \\ 
Min-p               & 73.3 & 75.6 & 76.6 & 35.3 & 36.2 & 36.9 & 45.5 & 47.2 & 47.6 \\ 
EDT                 & \textbf{74.9} & 75.6 & 77.0 & 34.7 & 36.2 & 36.8 & 45.5 & \textbf{47.8} & 48.2 \\ 
$\eta$-sampling     & 69.0 & 72.4 & 74.2 & 33.6 & 34.8 & 35.7 & 41.1 & 43.6 & 45.3 \\ 
$\varepsilon$-sampling & 73.4 & 76.4 & 78.2 & 35.0 & 36.2 & 36.9 & 45.4 & 46.9 & 47.7 \\ 
\midrule
\textbf{Greedy-Threshold}    & 70.1 & 73.6 & 75.5 & 33.6 & 34.9 & 36.1 & 40.3 & 42.9 & 44.4 \\ 
\textbf{Calibrated-TopK}     & 72.3 & 75.2 & 76.3 & \textbf{35.6} & \textbf{36.4} & \textbf{37.0} & \textbf{45.8} & 47.7 & \textbf{48.5} \\ 
\textbf{Calibrated-}$\varepsilon$ & 74.3 & \textbf{77.2} & \textbf{78.4} & 34.8 & 35.9 & 36.9 & 46.2 & 47.2 & 48.2 \\ 
\bottomrule
\end{tabular}
}
\end{table*}

We further compare our proposed methods against other existing methods across different benchmarks. Calibrated-TopK has the strongest overall performance for Qwen2.5-1.5B-Instruct.

We extend our analysis of Greedy-Threshold for up to 32B parameters models, considering both instruct and non-instruct models. As expected, larger models with stronger baseline performance are more challenging to improve. Existing samplers provides little improvement on the baseline. Nevertheless, Greedy-Threshold does not degrade performance. It either provides modest gains or remains comparable to existing samplers. One explanation for this diminishing effect is that larger models produce high-confidence predictions more frequently (\cref{fig:model sizes}), leading to fewer low-confidence steps. Since Greedy-Threshold only intervenes under low-confidence conditions, its impact naturally diminishes as model size increases.

\begin{table}[!b]
\centering
\caption{Majority voted results on GSM8K and Big-Bench-Hard. In addition to existing sampling conditions, Greedy-Threshold $p_{GT}=0.3$ is applied and shows strong consistent gains in addition to base samplers. Statistically significant differences ($p<0.05$) marked in \textbf{bold}.}
\label{}
\begingroup
\setlength{\tabcolsep}{2pt}       
\renewcommand{\arraystretch}{1.14}
{\footnotesize
\begin{tabular}{lcccccccc}
\toprule
\multirow{2}{*}{Method}
& \multicolumn{4}{c}{GSM8K}
& \multicolumn{4}{c}{Big-Bench-Hard} \\
\cmidrule(lr){2-5}\cmidrule(lr){6-9}
& {\small maj@1} & {\small maj@8} & {\small maj@16} & {\small maj@32} & {\small maj@1} & {\small maj@8} & {\small maj@16} & {\small maj@32} \\
\midrule
Baseline $T{=}1$ 
& 17.3 & 30.2 & 35.2 & 38.6 & 17.4 & 17.9 & 20.0 & 16.2   \\
\quad + Greedy-Threshold 
& +0.3 & +1.0 & +1.8 & +2.0 & +1.9 & \textbf{+2.9} & \textbf{+0.3} & \textbf{+3.2}  \\
\cmidrule(lr){1-9}
top-$k$ 
& 18.8 & 32.6 & 38.7 & 41.9 & 20.5 & 22.0 & 21.7 & 21.5 \\
\quad + Greedy-Threshold 
& +0.2 & \textbf{+1.1} & +0.5 & +1.1 & +1.1 & \textbf{+1.6} & \textbf{+1.5} & \textbf{+1.5}  \\
\cmidrule(lr){1-9}
top-$p$
& 22.4 & 35.5 & 40.8 & 43.6 & 22.3 & 25.5 & 25.8 & 25.9 \\
\quad + Greedy-Threshold 
& +0.6 & +0.9 & +0.4 & +1.3 & +1.6 & \textbf{+1.5} & \textbf{+1.5} & \textbf{+1.9} \\
\cmidrule(lr){1-9}
min-$p$
& 25.3 & 38.7 & 43.1 & 46.6 & 27.5 & 30.6 & 31.5 & 31.7\\
\quad + Greedy-Threshold 
& +1.9 & +1.4 & +0.8 & +0.6 & +0.3 & +0.2 & +0.1 & +0.1 \\
\cmidrule(lr){1-9}
EDT
& 28.0 & 40.2 & 44.7 & 46.8 & 27.0 & 30.4 & 31.1 & 31.7 \\
\quad + Greedy-Threshold
& 0.0 & +0.2 & -0.3 & +0.1 & +0.5 & +0.3 & +0.5 & \textbf{+0.3}  \\
\cmidrule(lr){1-9}
$\eta$-sampling
& 19.0 & 31.6 & 37.2 & 41.0 & 19.7 & 20.6 & 20.3 & 19.6 \\
\quad + Greedy-Threshold 
& +4.3 & \textbf{+2.4} & \textbf{+1.8} & \textbf{+1.7} & \textbf{+0.9} & \textbf{+1.9} & \textbf{+2.2} & \textbf{+2.7} \\
\bottomrule
\end{tabular}
}
\endgroup
\end{table}

\begin{table}[t]
\centering
\caption{Majority voted results on GSM8K. Greedy-Threshold improves performance more in smaller models. Statistically significant difference ($p<0.05$) marked in \textbf{bold}.}.
\label{app:tab:32b}
\begingroup
\setlength{\tabcolsep}{2pt}       
\renewcommand{\arraystretch}{1.14}
{\footnotesize
\begin{tabular}{lccccccccc}
\toprule
\multirow{2}{*}{Method}
& \multicolumn{3}{c}{Llama-3.2-1B}
& \multicolumn{3}{c}{Qwen2.5-14B-Instruct}
& \multicolumn{3}{c}{Qwen2.5-32B-Instruct} \\
\cmidrule(lr){2-4}\cmidrule(lr){5-7}\cmidrule(lr){8-10}
& {\small maj@8} & {\small maj@16} & {\small maj@32} & {\small maj@8} & {\small maj@16} & {\small maj@32} & {\small maj@8} & {\small maj@16} & {\small maj@32} \\
\midrule
Baseline $T{=}1$ 
& 0.7 & 0.6 & 0.2 & 92.8 & 93.4 & 93.5 & 92.8 & 93.4 & 94.0 \\
\quad + Greedy-Threshold 
& +0.7 & \textbf{+0.5} & +0.5 & +0.2 & +0.2 & \textbf{+0.3} & +0.1 & +0.3 & -0.2 \\
\cmidrule(lr){1-10}
Top-$k$ 
& 2.6 & 1.7 & 1.4 & 93.3 & 93.5 & 93.7 & 92.9 & 93.7 & 93.9 \\
\quad + Greedy-Threshold 
& -0.3 & +0.6 & +0.1 & 0.0 & -0.1 & 0.0 & +0.1 & -0.1 & 0.0 \\
\cmidrule(lr){1-10}
Top-$p$ 
& 1.7 & 1.4 & 1.6 & 93.3 & 93.5 & 93.5 & 93.1 & 93.5 & 93.7 \\
\quad + Greedy-Threshold 
& +0.3 & +0.5 & +0.4 & +0.1 & -0.1 & 0.0 & -0.2 & -0.1 & -0.2 \\
\cmidrule(lr){1-10}
Min-$p$ 
& 3.9 & 3.9 & 3.6 & 93.2 & 93.1 & 93.3 & 92.8 & 93.4 & 93.6 \\
\quad + Greedy-Threshold 
& +0.9 & +0.5 & +0.9 & -0.3 & +0.2 & +0.1 & +0.3 & 0.0 & -0.2 \\
\cmidrule(lr){1-10}
EDT 
& 4.2 & 3.8 & 3.9 & 92.9 & 93.3 & 93.4 & 92.7 & 93.3 & 93.5 \\
\quad + Greedy-Threshold 
& +0.9 & +0.3 & +0.1 & +0.1 & -0.1 & 0.0 & -0.1 & -0.1 & +0.1 \\
\cmidrule(lr){1-10}
$\eta$-sampling 
& 1.1 & 0.8 & 0.5 & 93.5 & 93.6 & 93.7 & 93.2 & 93.6 & 93.9 \\
\quad + Greedy-Threshold 
& +0.7 & +0.5 & +0.7 & -0.4 & -0.3 & +0.1 & -0.1 & 0.0 & -0.1 \\
\bottomrule
\end{tabular}
}
\endgroup
\end{table}

\newpage

\subsection{Effectiveness at Low Temperatures}\label{app:effectiveness at low temperatures}
\begin{table}[!b]
\centering
\caption{Majority voted results on GSM8K with scaled temperature $T=0.6$, $\varepsilon=0.01$, $p_{GT}=0.1$ and $c_{CT}=0.01$}.
\label{app:tab:t=0.6}
\begingroup
\setlength{\tabcolsep}{2pt}       
\renewcommand{\arraystretch}{1.14}
{\footnotesize
\begin{tabular}{lcccccc}
\toprule
\multirow{2}{*}{Method}
& \multicolumn{3}{c}{{Qwen2.5-0.5B-Instruct}}
& \multicolumn{3}{c}{Qwen2.5-1.5B-Instruct} \\
\cmidrule(lr){2-4}\cmidrule(lr){5-7}
& {\small maj@8} & {\small maj@16} & {\small maj@32} & {\small maj@8} & {\small maj@16} & {\small maj@32} \\
\midrule
No conditions 
& 40.7 & 44.9 & 46.9 & 73.7 & 76.3 & 76.9  \\
$\varepsilon$-sampling
& \textbf{41.4} & \textbf{45.2} & 46.8 & \textbf{74.4} & 75.8 & 77.0  \\
\cmidrule(lr){1-7}
\textbf{Greedy-Threshold} 
& 41.2 & \textbf{45.2} & 47.2 & 74.3 & \textbf{76.9} & 77.1  \\
\textbf{Calibrated-TopK}
& 40.9 & 44.5 & 46.7 & 74.1 & 76.0 & \textbf{77.3}  \\
\textbf{Calibrated}-$\varepsilon$
& \textbf{41.4} & \textbf{45.2} & \textbf{48.4} & 74.2 & 75.6 & 77.0  \\
\bottomrule
\end{tabular}
}
\endgroup
\end{table}

Since lower temperatures are used for math and coding tasks, we test our proposed methods using $T=0.6$. The reason for temperature scaling is to make probability distribution more peaked so that top-1 probability increases while the tail probabilities shrink. Consequently, probability-based pruning becomes implicitly more aggressive. Many low-rank tokens become unlikely to be sampled even without changing any thresholds. This aligns with our thesis that low ranked tokens should not be broadly sampled due to their correlation with low correctness. Given the same non top-1 ranked token, after temperature scaling, its probability will decrease. Thus, to remove the same token as without temperature scaling, the probability cutoff needs to be lower. We choose $\varepsilon=0.01$, $p_{GT}=0.1$ and $c_{CT}=0.01$. 

Given the same $p_{max}$, its scaled probability would be bigger. Thus, we should set a higher Greedy-Threshold than before. However, in practice we found that this limits diversity significantly that the benefit from self-consistency diminishes. The maj@1 accuracy increases but performance gain from maj@k reduces. We hypothesize that this is because more lower temperature already results in diminished diversity. Further restrictions results in diversity collapse. From \cref{app:tab:t=0.6}, we can see that further gains from baseline is much smaller than with $T=1$. Nevertheless, by setting lower truncation thresholds, we still see performance gains from using our proposed truncation methods.

\newpage

\subsection{What About Creative Writing?}
Our main experiments target reasoning tasks with closed-form answers. A natural question is whether the same correctness-aware perspective applies to open-ended generation. We provide an illustrative case study on creative writing using LitBench \citep{fein_litbench_2025}, treating the \emph{chosen} story as a proxy for ground truth. While we do not evaluate samplers on creative metrics here, we construct calibration grids and probability–correctness scatter plots and observe qualitatively similar patterns. As confidence decreases, expected correctness declines. As rank increases, correctness drops sharply. Compared with GSM8K and Alpaca calibration, creative prompts exhibit systematically lower confidence, with low-confidence bins occurring more often (e.g., lowest-bin frequency: \SI{6.01}{\percent} for LitBench vs.\ \SI{0.17}{\percent} for Alpaca and \SI{0}{\percent} for GSM8K). The probability–correctness mapping in log–log space is also stronger in this setting, with a slope closer to one. Full calibration diagrams and scatter plots are provided in \cref{app:Example Calibration and Linear Fit Diagrams}. We refrain from drawing conclusions about the effectiveness of our proposed samplers for creative writing, but highlight these trends as an intriguing observation and a direction for future work.

\subsection{Example Calibration Diagrams}\label{app:Example Calibration and Linear Fit Diagrams}

\begin{figure}[htbp]
    \centering
    \begin{subfigure}[b]{0.325\textwidth}
        \centering
        \includegraphics[width=\linewidth]{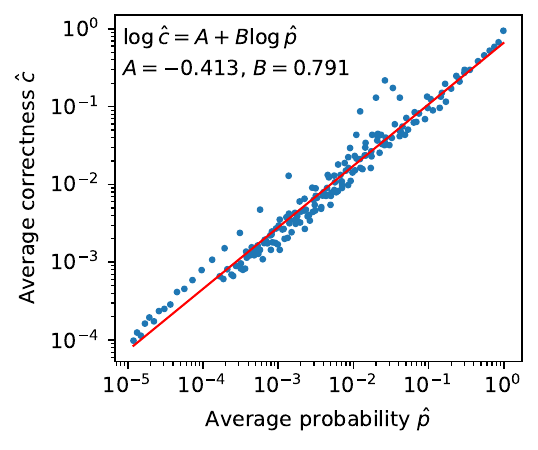}
        \caption{GSM8K}
    \end{subfigure}
    \hfill
    \begin{subfigure}[b]{0.33\textwidth}
        \centering
        \includegraphics[width=\linewidth]{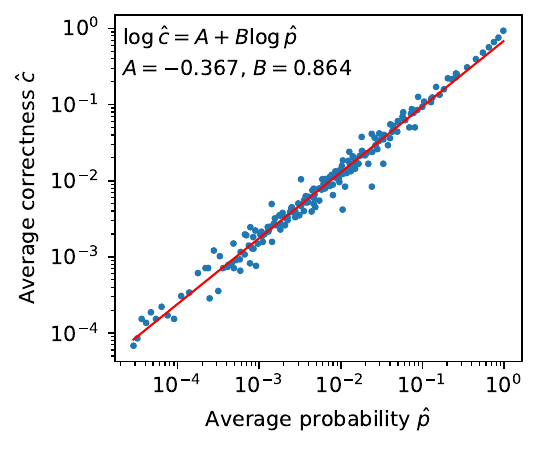}
        \caption{alpaca-gpt4-en}
    \end{subfigure}
    \hfill
    \begin{subfigure}[b]{0.33\textwidth}
        \centering
        \includegraphics[width=\linewidth]{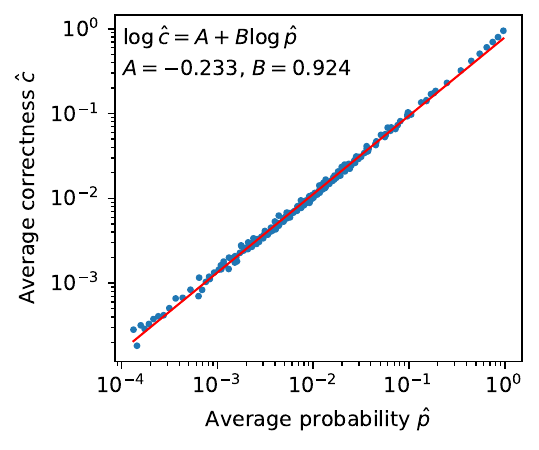}
        \caption{LitBench}
    \end{subfigure}
    \caption{Calibration scatter plots on Qwen2.5-1.5B-Instruct}
\end{figure}

\begin{figure}[htbp]
    \centering
    \begin{subfigure}[b]{0.44\textwidth}
        \centering
        \includegraphics[width=\linewidth]{qwen2.5-0.5b-ins-gsm8k-scatter.pdf}
        \caption{GSM8K}
    \end{subfigure}
    \begin{subfigure}[b]{0.44\textwidth}
        \centering
        \includegraphics[width=\linewidth]{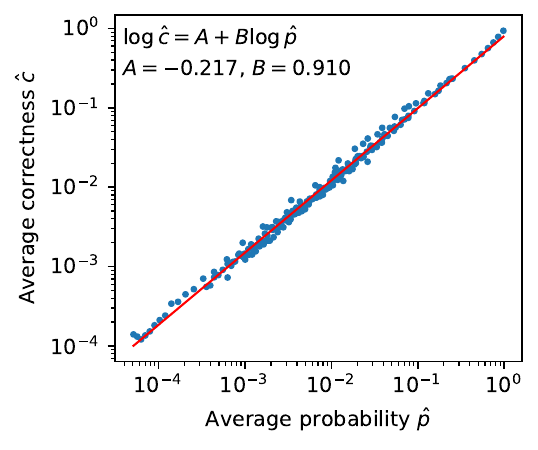}
        \caption{alpaca-gpt4-en}
    \end{subfigure}
    \caption{Calibration scatter plots on Qwen2.5-0.5B-Instruct}
\end{figure}

\begin{figure}[p]
    \centering
    \begin{subfigure}{\textwidth}
        \centering
        \includegraphics[width=0.9\textwidth]{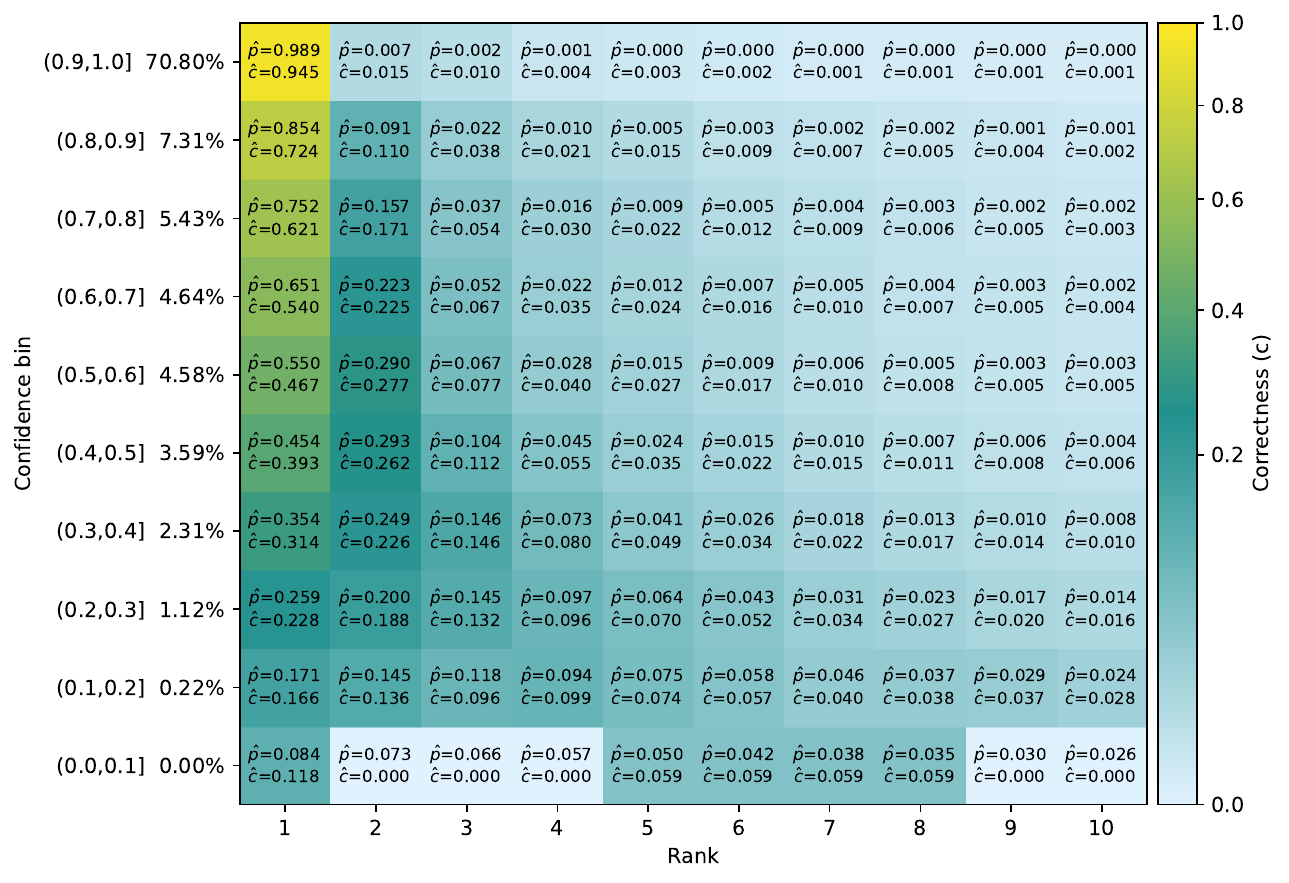} 
        \caption{Qwen2.5-0.5B-Instruct on GSM8K}
    \end{subfigure}
    \begin{subfigure}{\textwidth}
        \centering
        \includegraphics[width=0.9\textwidth]{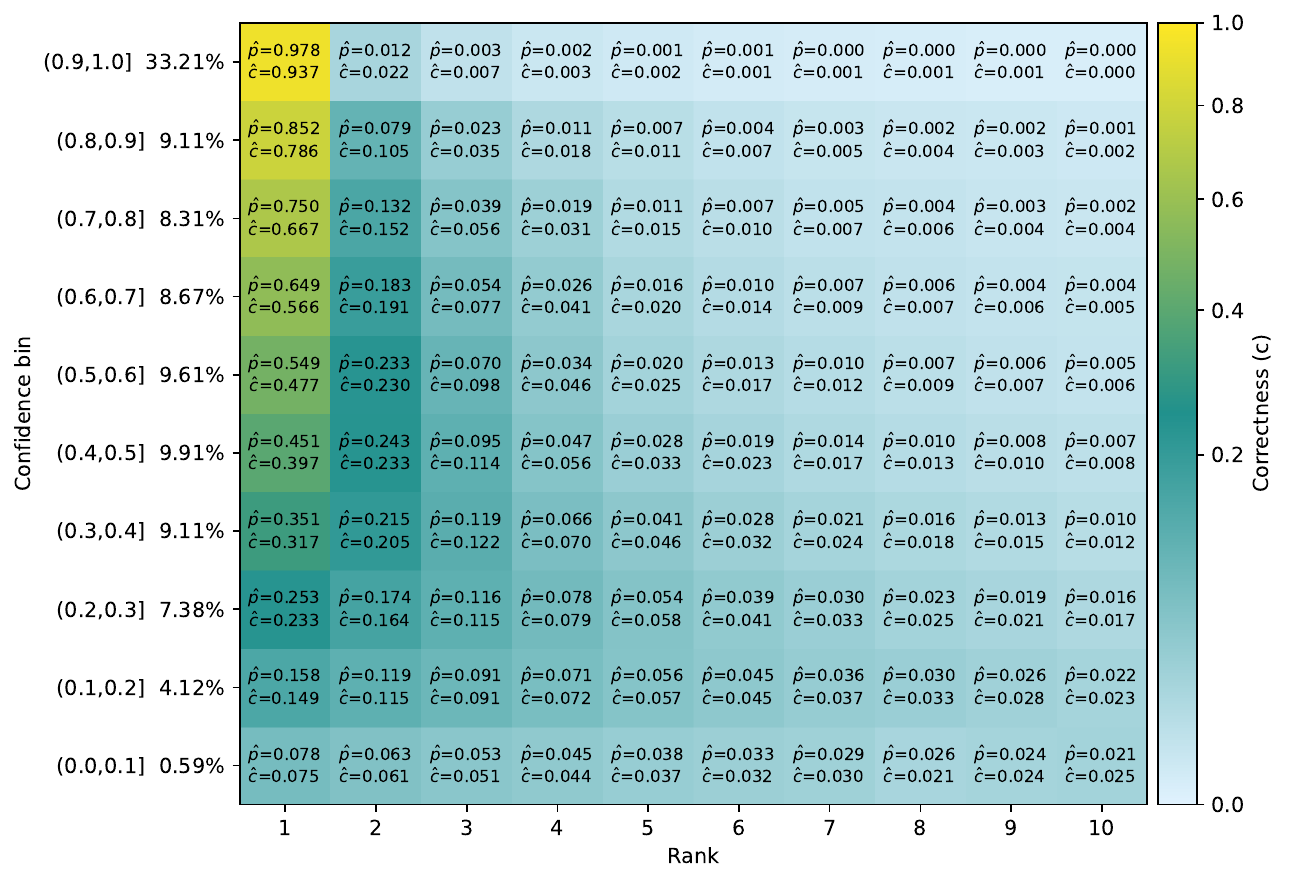} 
        \caption{Qwen2.5-0.5B-Instruct on alpaca-gpt4-en}
        \label{subfig:second}
    \end{subfigure}
    \caption{Calibration grids on various Qwen models and GSM8K or alpaca-gpt4-en}
\end{figure}

\begin{figure}[p]
    \centering
    \ContinuedFloat
    \begin{subfigure}{\textwidth}
        \centering
        \includegraphics[width=0.9\textwidth]{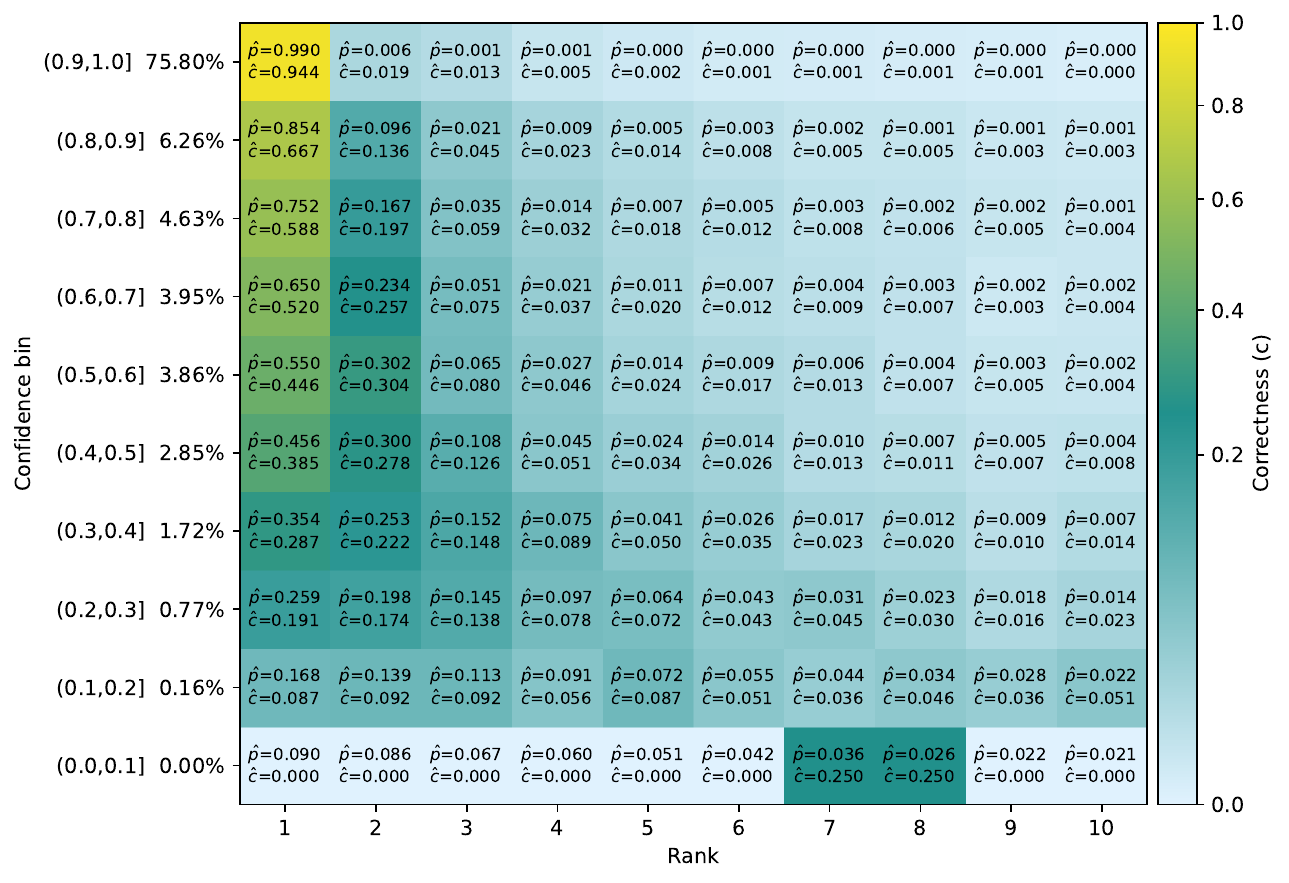}
        \caption{Qwen2.5-1.5B-Instruct on GSM8K}
    \end{subfigure}
    \begin{subfigure}{\textwidth}
        \centering
        \includegraphics[width=0.9\textwidth]{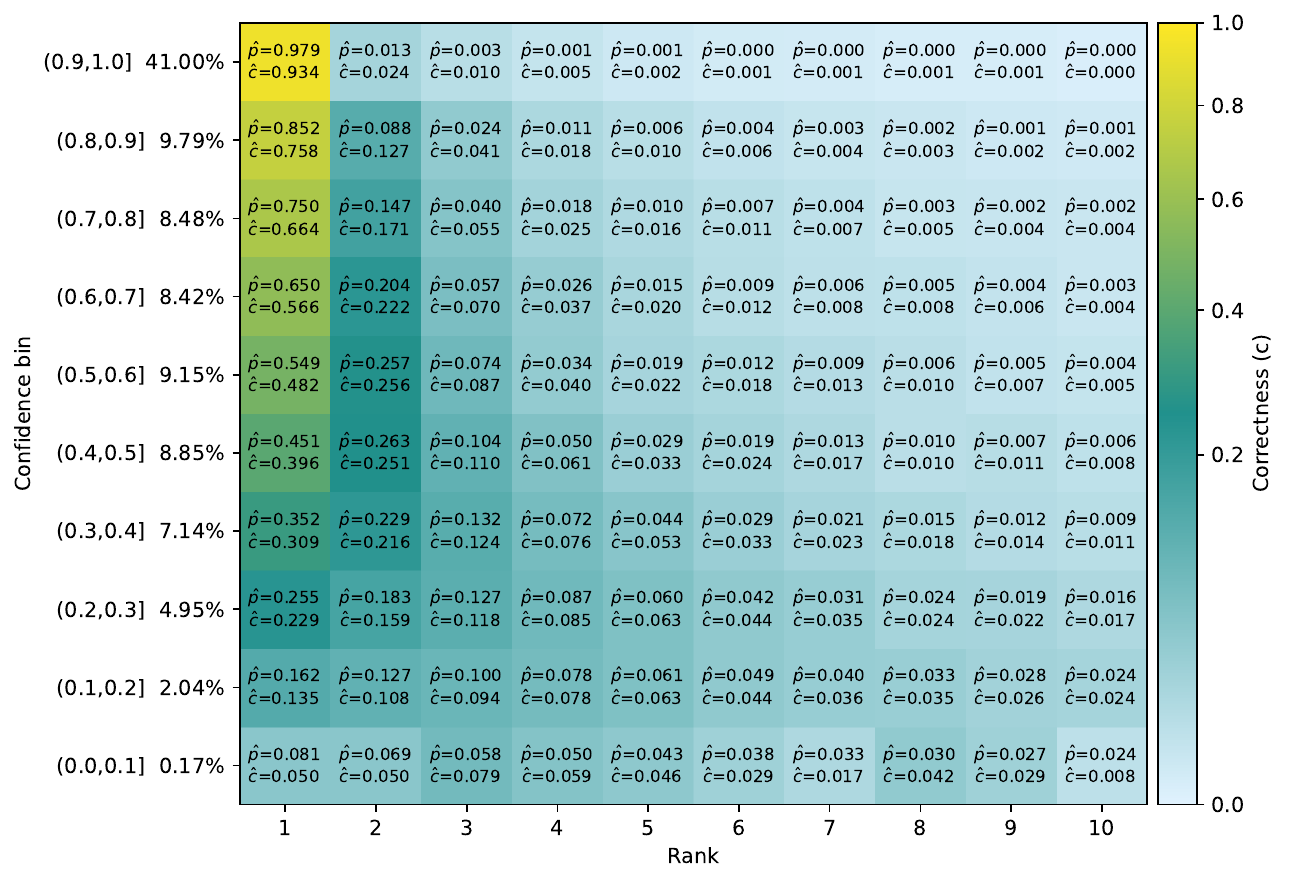} 
        \caption{Qwen2.5-1.5B-Instruct on alpaca-gpt4-en}
    \end{subfigure}
    \caption{Calibration grids on various Qwen models and GSM8K or alpaca-gpt4-en}
\end{figure}

\begin{figure}[p]
    \centering
    \ContinuedFloat
    \begin{subfigure}{\textwidth}
        \centering
        \includegraphics[width=0.9\textwidth]{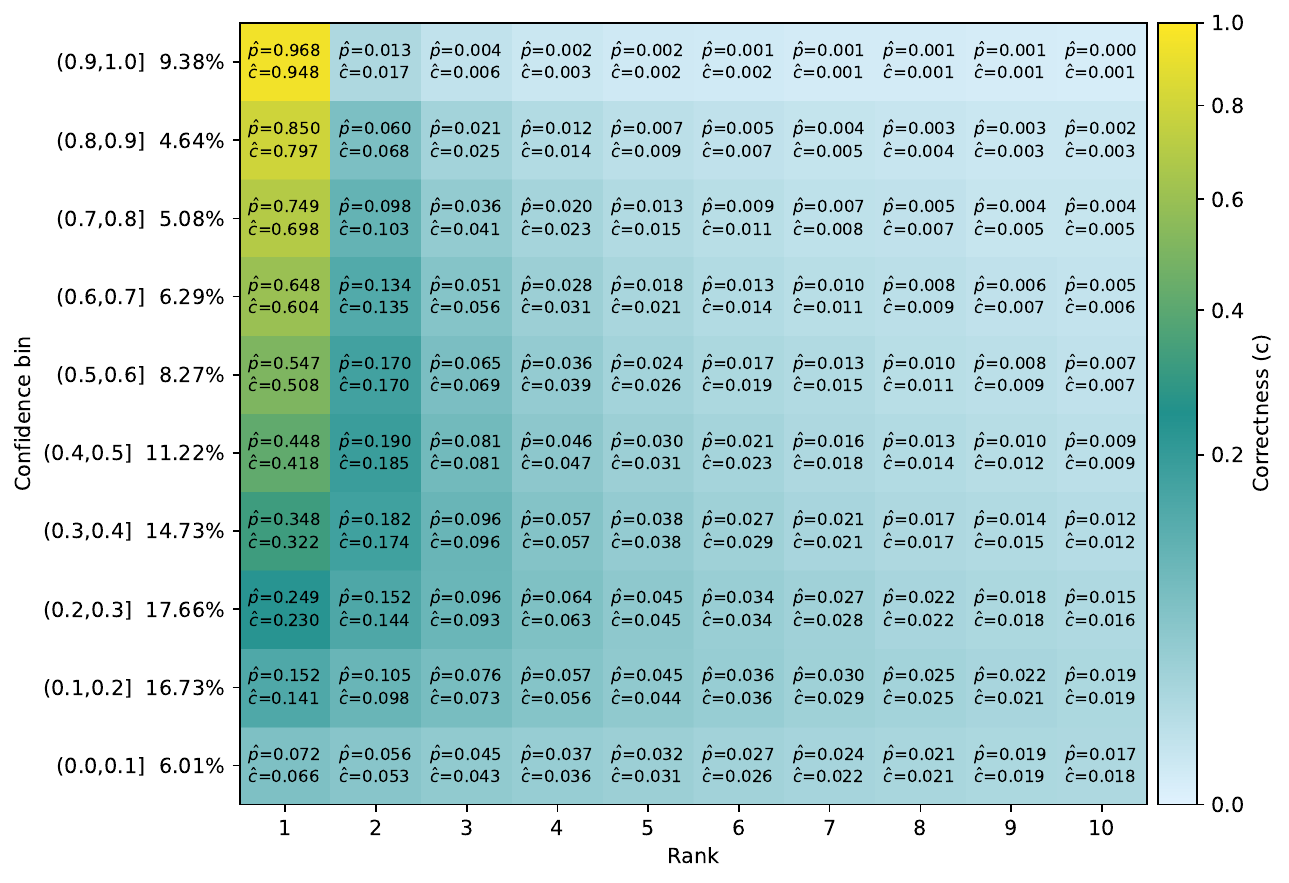}
        \caption{Qwen2.5-1.5B-Instruct on LitBench}
    \end{subfigure}
    \caption{Calibration grids on various Qwen models and GSM8K, alpaca-gpt4-en or LitBench}
\end{figure}

\end{document}